\documentclass[a4paper,fleqn]{cas-dc}

\usepackage[authoryear,round]{natbib}

\usepackage{amsmath,amssymb}
\usepackage{algorithm}
\usepackage{algpseudocode}
\usepackage{booktabs}
\usepackage{multirow}
\usepackage{makecell}
\usepackage{changepage}
\usepackage{adjustbox}
\newcommand{\sur}[1]{#1}
\usepackage{graphicx}
\usepackage{xcolor}
\usepackage[switch]{lineno}

\usepackage[normalem]{ulem}

\begin{document}

\shorttitle{Dynamic gain neuromodulation attenuates stability gaps during continual joint training}    

\shortauthors{Rodriguez-Garcia et al.}  

\title [mode = title]{Dynamic gain neuromodulation attenuates stability gaps during continual joint training}  



\author[1]{Alejandro \sur{Rodriguez-Garcia}}[orcid=0000-0002-7230-4156]
\ead{a.rodriguez-garcia2@newcastle.ac.uk} 
\credit{
Conceptualization,
Formal analysis,
Methodology,
Software,
Validation,
Investigation,
Visualization,
Writing -- original draft,
Writing -- review and editing}

\affiliation[1]{
    organization={Neural Circuits Laboratory, Biosciences Institute, Newcastle University},
    city={Newcastle upon Tyne},
    postcode={NE2 4HH},
    country={United Kingdom}
}

\author[1]{Anindya Ghosh}
\credit{Supervision, Writing -- review and editing}

\author[1]{Srikanth Ramaswamy}
\cormark[1]
\ead{srikanth.ramaswamy@newcastle.ac.uk}
\credit{ 
Supervision, 
Resources, 
Funding acquisition, 
Writing -- review and editing}

\cortext[1]{Corresponding author}



\begin{abstract}
Recent work in continual learning has highlighted the \emph{stability gap} -- a temporary performance drop on previously learned tasks when new ones are introduced. This phenomenon reflects a mismatch between rapid adaptation and strong retention at task boundaries, underscoring the need for optimization mechanisms that balance plasticity and stability over abrupt distribution changes. While optimizers such as momentum-SGD and Adam introduce implicit multi-timescale behavior, they still exhibit pronounced stability gaps. Importantly, these gaps persist even under ideal joint training, making it crucial to study them in this setting to isolate their causes from other sources of forgetting. Motivated by how noradrenergic (neuromodulatory) bursts transiently increase neuronal gain under uncertainty, we introduce a dynamic gain scaling mechanism as a two-timescale optimization technique that balances adaptation and retention by transiently increasing the effective update magnitude while dynamically reparameterizing the weights governing the forward pass, thereby empirically mitigating transition-induced curvature amplification. Across domain- and class-incremental MNIST, CIFAR, and mini-ImageNet benchmarks under task-agnostic joint training, dynamic gain scaling effectively attenuates stability gaps while maintaining competitive accuracy, improving robustness at task transitions.
\end{abstract}




\begin{NoHyper}
\maketitle
\end{NoHyper}


\section{Introduction}\label{sec:1}
\begin{figure}[pos=!htbp]
  \centering
    \includegraphics[width=0.9\columnwidth]{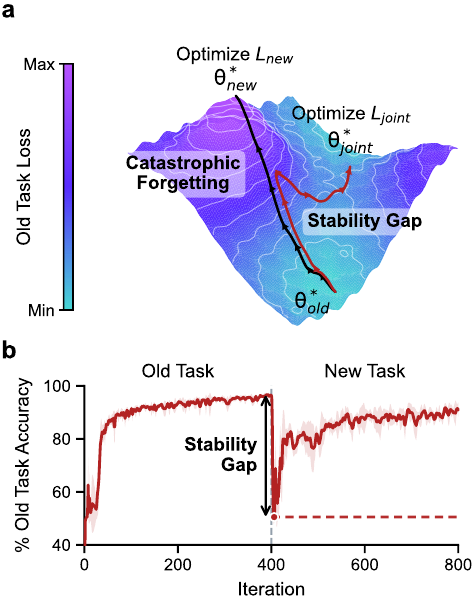}  
    \caption{\textbf{Schematic of the stability gap.} \textbf{a)} Conceptual illustration highlighting the distinction between the stability gap and catastrophic forgetting. While catastrophic forgetting arises from an inaccurate approximation of the joint loss, the stability gap emerges from a deviation of the optimization trajectory from the path of non- or minimally-increasing loss, leading to a transient drop in old-task performance. \textbf{b)} Old-task accuracy during sequential learning on Split CIFAR-10 under stochastic gradient descent with 0.9 momentum. The sharp performance drop at the task boundary quantifies the stability gap that arises from the deviated trajectory.}  \label{fig:stability_gap_explainer}
\end{figure}

Continual learning (CL) in artificial neural networks aims to emulate lifelong adaptability observed in biology, where networks continuously acquire new data without \emph{catastrophically} overwriting knowledge from past examples~\citep{van_de_ven_three_2022, kudithipudi_biological_2022, mei_improving_2025}. Traditionally, most work in this field has concentrated on mitigating this phenomenon by approximating the ideal joint loss over all previous tasks encountered~\citep{van_de_ven_three_2022, de_lange_continual_2022}. However, recent research has identified the presence of \emph{stability gaps}~\citep{de_lange_continual_2022, hess_two_2023} -- a transient forgetting of past task knowledge at every task switch (Figure~\ref{fig:stability_gap_explainer}). Crucially, this gap remains even under ideal joint training, indicating that it originates from the \emph{dynamics of optimization} rather than from representational drift or limited memory capacity~\citep{hess_two_2023, kamath_expanding_2024} and runs counter to the core objective of stable CL systems. Practically, although a model \emph{may} recover some performance after an initial decline, this transient drop in accuracy introduces significant risk when updating models frequently~\citep{de_lange_continual_2022}. As a result, models tend to be updated less often or are retrained entirely from scratch, which is computationally expensive.

Standard optimizers such as momentum-SGD (MSGD) \citep{rumelhart_learning_1986} and Adam~\citep{kingma_adam_2014} introduce implicit multi-timescale regularization through momentum or adaptive learning rates, yet they still exhibit stability gaps. Recent analyses show that momentum terms can deviate the update from the true gradient direction~\citep{kamath_expanding_2024}, causing overshooting when the loss landscape shifts abruptly at task boundaries. We therefore view the stability gap as an early-adaptation manifestation of the stability-plasticity dilemma, arising when rapid acquisition of new information transiently perturbs previously acquired knowledge. This makes the stability gap a dynamics-induced issue that remains unresolved, and one that is particularly detrimental in \emph{online} CL, where models may be queried during deployment and must therefore remain robust throughout training.

However, in biology, neural circuits demonstrate lifelong adaptability, continually integrating new information with minimal disruption to established memories~\citep{kudithipudi_biological_2022}. This finely tuned balance between plasticity and stability arises from intrinsic neuromodulatory dynamics in cortical networks~\citep{mei_improving_2025, shine_neuromodulatory_2019}. A key mechanism supporting this flexibility is gain neuromodulation -- transient changes in neuronal responsiveness that adjust how strongly inputs influence neural activity without altering selectivity~\citep{ferguson_mechanisms_2020, jordan_locus_2024}. Although gain modulation can arise through multiple biological pathways, a well-characterized source is noradrenergic activity that increases neural gain in response to unexpected uncertainty during sensory evidence integration, enabling rapid updates to evolving perceptual estimates while preserving stable representations~\citep{aston-jones_integrative_2005, shine_neuromodulatory_2019, munn_ascending_2021, wainstein_gain_2025}.

We hypothesize that these transient gain boosts induce fast and slow timescales of weight adaptation, enabling rapid adjustments to new information while preserving stable representations, analogous to the fast-slow weight optimization paradigm -- where fast weights momentarily improve performance on earlier tasks during new learning without disrupting the consolidation occurring in slow weights~\citep{hinton_using_1987}. In addition, gain modulation induces a dynamic reparameterization between the base weights updated by gradient descent and the effective weights expressed in the forward pass \citep{dinh_bengio_icml17}. We hypothesize that this forward-pass reparameterization attenuates the transition-induced curvature amplification and loss sensitivity that would otherwise accompany the larger effective update.

Since efforts to understand the stability gap have led to the realization that we should focus not only on \emph{what} to optimize, but also on \emph{how} to optimize (i.e. the optimization technique~\citep{hess_two_2023}, see Figure~\ref{fig:stability_gap_explainer}), we introduce an uncertainty-modulated gain mechanism to act as an approximate two-timescale gradient descent adjustment to mitigate stability gaps in \emph{task-agnostic online} CL scenarios. We benchmark against Adam~\citep{kingma_adam_2014} and MSGD~\citep{rumelhart_learning_1986}, as both are widely used multi-timescale optimizers, and also include vanilla SGD~\citep{goodfellow_DL2016} as a non-adaptive single-timescale baseline. Thus, the main contributions of our work are:
\begin{itemize}
    \item We introduce uncertainty-guided gain dynamics and analytically show that transient gain boosts in gradient descent updates induce emergent fast and slow weight-adaptation timescales, mirroring multi-timescale optimization methods.
    \item We derive how neuronal gain transforms gradients and curvature between the base weights updated by gradient descent and the effective weights expressed in the forward pass and empirically show that this attenuates task transition-induced curvature amplifications associated with rapid adaptation.
    \item Building on these insights, we introduce NGM-SGD, an optimizer for \emph{online} continual learning that empirically attenuates the stability gap across domain- and class-incremental MNIST, CIFAR, and mini-ImageNet benchmarks under task-agnostic joint training, outperforming MSGD, Adam, and vanilla SGD in transition stability.
\end{itemize}
Note that as our contribution concerns \emph{how} optimization is performed rather than \emph{what} continual learning strategy is employed, NGM-SGD is complementary to existing approaches for catastrophic forgetting, such as replay methods (see Appendix~\ref{app:replay}) and regularization techniques. Although evaluated here under idealized joint training to isolate the stability gap, it can be combined directly with these methods.
\subsection{Biological basis for noradrenergic gain}
Incremental learning tasks in AI systems~\citep{van_de_ven_three_2022}, where networks must adapt internal representations in response to changing data distributions, can be viewed as a form of \emph{perceptual updating} in neuroscience experiments -- the process by which new sensory information is acquired sequentially by animals. In biological systems, such updating is not solely mediated by synaptic plasticity but is also coordinated by neuromodulatory signals that regulate when internal representations should be revised. Specifically, noradrenergic (NA) bursts from the locus coeruleus (LC) -- crucial for regulating arousal, stress responses, attention, and sleep-wake cycles -- play a vital role in these behaviors by promoting knowledge acquisition under uncertainty~\citep{sales2019}.

Experimental and theoretical studies have shown that, mechanistically, phasic LC-NA bursts -- conveying urgent behavioral demands such as novel stimuli -- transiently increase neuronal gain, globally amplifying neural responsiveness without altering selectivity~\citep{ferguson_mechanisms_2020, jordan_locus_2024}. This gain increase effectively flattens the underlying neural-state energy landscape, enabling rapid transitions between competing representations and facilitating perceptual switches~\citep{wainstein_gain_2025, munn_neuronal_2023}. In particular, \citet{wainstein_gain_2025} demonstrated that under perceptual ambiguity, pupil diameter -- an indirect and non-specific readout of phasic LC activity and neuromodulatory tone~\citep{samuels_szabadi_lc_2008} -- is tightly linked to switches in perceptual content, as pupil-linked gain fluctuations recapitulate the phasic-tonic firing patterns characteristic of LC dynamics and act as a transient ``network reset'', facilitating transitions in perceptual categorization. Thus, interestingly, evoked pupil dilations reliably coincide with the resolution of perceptual ambiguity, supporting the view that uncertainty-driven NA gain modulation plays a causal role in perceptual updating \citep{sales2019}.

Based on these insights (see Appendix~\ref{app:gain_dynamics}, where we show that our gain dynamics reduce to the formulation of~\citet{wainstein_gain_2025}), we hypothesize that these transient gain boosts on neural circuits also induce fast and slow timescales of weight adaptations that facilitate knowledge acquisition, while simultaneously driving a reparameterization that generates fast yet stable transitions between tasks~\citep{hinton_using_1987, dinh_bengio_icml17}. Hence, we apply this biologically inspired idea to learning dynamics in artificial networks, asking how uncertainty-driven gain modulation can shape optimization trajectories during training. However, while in biological circuits, gain modulation arises through spiking and population-level dynamics~\citep{munn_neuronal_2023, whyte_burst-dependent_2023}, in artificial networks, we implement it by scaling incoming weights with a gain coefficient under a Rectified Linear Unit (ReLU) activation function. Thus, this work constitutes a minimal and biologically interpretable abstraction that preserves the functional role of noradrenergic gain as a modulator of input sensitivity, enabling transient reshaping of the effective-weight loss landscape and adaptive updating of network representations.

\subsection{Related work}\label{sec:related}
\paragraph{Stability gaps in CL.} It was not until recently that a study pointed out the conventional evaluation paradigm for AI systems is fundamentally inadequate~\citep{de_lange_continual_2022, hess_two_2023}. Consider a sequence of tasks $\{T_k\}_{k=1}^K$  with corresponding data distributions $\{D_k\}_{k=1}^K$ and let $\mathcal{F}_\theta$ denote a model parameterized by~$\theta$. Under the traditional task-based evaluation, $\mathcal{F_\theta}$ is trained sequentially, updating its parameters $\theta_k$ upon completing task $T_k$, and its performance is measured only immediately after training on $T_k$ has finished~\citep{de_lange_continual_2022}. Notably, this task‐based evaluation is blind to how a model’s accuracy evolves during training. As an alternative, the continual evaluation strategy measures the performance of $\mathcal{F_\theta}$ every $\rho$ training iterations while learning task $T_k$~\citep{de_lange_continual_2022}. Interestingly, continual evaluation revealed stability gaps~\citep{de_lange_continual_2022, caccia_new_2021} even in large language models~\citep {guo_efficient_2024}. These brief accuracy drops undermine the goals of CL by introducing temporary forgetting between tasks and limiting the system’s overall learning capacity. Specifically, this might be especially problematic in continual safety-critical applications, where even momentary failures can have severe consequences~\citep{de_lange_continual_2022, hess_two_2023}. One may naively attribute the stability gap to imperfect approximations of the joint objective by CL methods. Yet this hypothesis does not hold up as the phenomenon persists even under perfect joint training, where $L_{\text{joint}} = L_{\text{new}} + {L}_{\text{old}}$~\citep{hess_two_2023, kamath_expanding_2024}. Hence, this finding implies that the transient performance collapse is not merely an artifact of memory or regularization approximations, but rather emerges from the fundamental dynamics of sequential optimization, underscoring the importance of studying it in the idealized joint-loss setting~\citep{kamath_expanding_2024}.

Beyond evaluation paradigms and identification of the stability gap, a growing body of work has sought to pinpoint its causes and propose remedies. \citet{hess_two_2023} argue that there exists a non-increasing-loss path between $\theta_{\text{old}}$ and $\theta_{\text{joint}}$, suggesting that a continual learner only needs to follow this valley. However, enforcing such trajectories via gradient-projection methods has shown limited practical success. In a homogeneous CIFAR setting, \citet{kamath_expanding_2024} further demonstrate that although a linear low-loss path does exist, MSGD consistently deviates from it at each task transition. Although successful approaches to mitigate the stability gap exist, many rely on added architectural modules or task-specific details, making them impractical for lightweight, task-agnostic online use. For example, \citet{lapacz_exploring_2024} enlarge the classifier with each new task -- leading to growing model complexity -- and demonstrate that much of the transient forgetting originates in the final linear layer. \citet{harun_overcoming_2023} and \citet{guo_efficient_2024} rely on pretrained backbones and low-rank updates, and \citet{soutifcormerais_improving_2023} maintain a smoothed average of the weights across task transitions, improving stability at computational cost. These limitations highlight the need for lightweight, optimizer-level interventions compatible with task-agnostic online training. Our method takes precisely this route, introducing a bio-inspired gain-modulated update rule to mitigate the stability gap through the update dynamics.
\paragraph{Neuronal gain modulation.} Neuronal gain characterizes the input sensitivity of a neuron. It is formally defined as the slope of its input-output transfer function, reflecting the rate at which output firing rates change in response to variations in input current~\citep{ferguson_mechanisms_2020, jordan_locus_2024, aston-jones_integrative_2005}. This property has been extensively investigated in biophysical models of cortical circuits, highlighting the role of gain modulation in adaptive computation~\citep{munn_neuronal_2023, bos_untangling_2025, rodriguezgarcia2025rolegainneuromodulationlayer5}, and fine-tuning motor responses~\citep{stroud_motor_2018}. Furthermore, theoretical neuroscience works have debated the causal relationship between gain modulation and synaptic plasticity, suggesting a complex interplay~\citep{swinehart_supervised_2005}. However, despite its biological relevance, gain modulation has received limited attention in artificial learning systems. A recent study has used gain modulation for attentional gating in hierarchical models but only as a fine-tuning step \emph{after} weight learning, ignoring its interaction with weight updates~\citep{caroline_haimerl_fine-tuning_nodate}. Another study used it as an isolated mechanism for adaptive whitening~\citep{duong_adaptive_2023-1} and extended it to multi-timescale whitening by pairing fast gain changes with slower synaptic plasticity~\citep{duong_adaptive_2023}. However, their gain modulation was limited to interneurons in a single-layer architecture~\citep{duong_adaptive_2023-1, duong_adaptive_2023}. Nevertheless, a key theoretical insight is that gain neuromodulation reshapes the neural-state energy landscape: noradrenergic input flattens it, thus enabling flexible switching, while cholinergic input deepens it to stabilize memories~\citep{munn_ascending_2021, munn_neuronal_2023, shine_neuromodulatory_2019}. Recent recurrent neural network work supports this view, showing that uncertainty-driven gain modulation facilitates perceptual switches by propelling the system through high‑velocity trajectories~\citep{wainstein_gain_2025}. However, in that model, gains were fixed during training and modulated only afterwards, limiting interaction with weight updates. We extend this approach by including gain dynamics during learning and applying it to reduce stability gaps in joint CL.
\section{Results}
\subsection{Gain boosts as a multi-timescale optimizer}
Introducing processing timescales into artificial neural networks provides a principled strategy to reconcile stable, long-term memory consolidation with rapid, context-driven adaptation. The fast-slow weight paradigm exemplifies this principle by maintaining distinct ``slow'' and ``fast'' weight components, i.e. $W_{\textbf{eff}} = w_{\textbf{slow}} + w_{\textbf{fast}}$~\citep{hinton_using_1987}. This enables networks to swiftly adapt to distribution shifts via transient updates in the fast weights, which rapidly decay toward zero, while simultaneously preserving consolidated knowledge within slow weights~\citep{jones_learning_2022}. Crucially, this mechanism minimizes interference between newly acquired information and existing representations~\citep{hinton_using_1987}, making it particularly suitable as a foundation for lifelong learning in artificial systems.

\begin{figure*}[pos=!htbp]
  \centering
  \includegraphics[width=0.9\textwidth]{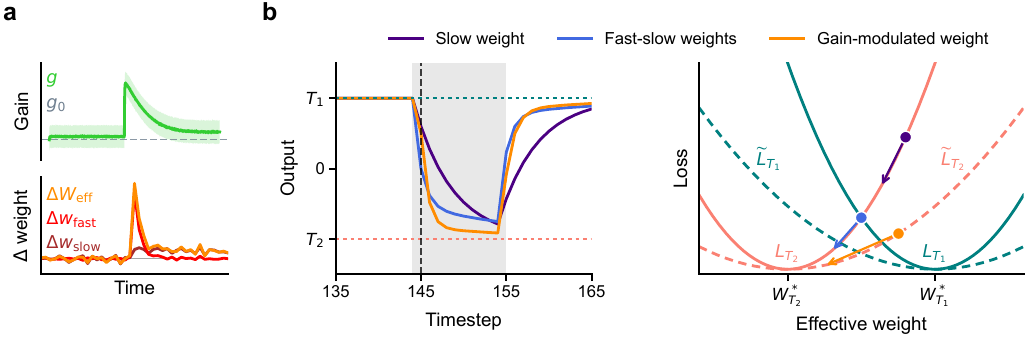}
\caption{\textbf{Simple proof of principle showing how gain boosts approximate a two-timescale optimizer.} \textbf{a)}~Schematic of gain-induced effective weight decomposition under noradrenergic neuromodulation. 
Phasic noradrenergic signals transiently increase neuronal gain, thereby ``virtually'' splitting the effective weight into a fast-slow scheme with $w_{\text{fast}} = \left[g(t) - g_{0}\right] w(t)$ and $w_{\text{slow}} = g_{0} \, w(t)$. \textbf{b)} \textit{Left panel.} Simplified model $y(t) = W_{\mathrm{eff}}(t)\,x$ with constant input $x$ predicting a target $T(t)$ under mean squared error loss $L = \tfrac{1}{2}[T(t)-y(t)]^2$. Comparison of gain-modulated (orange), fast-slow (blue) and slow (purple) weight methods. The unshaded region corresponds to optimization toward target $T_1$, and the shaded region to optimization toward target $T_2$. The dashed vertical line marks the time of peak neuronal gain. \textit{Right panel.} Effective-coordinate flattening induced by gain neuromodulation at the highest neuronal gain. In the effective-weight space, gain boosts reparameterize the loss, effectively reducing its curvature and scaling its gradient step. Dots illustrate each method's state at the peak gain time, and arrow lengths indicate the gradient-step magnitudes.}
  \label{fig:multitimescale-weight}
\end{figure*}

In biological neural circuits, analogous multi-timescale processing occurs through neuromodulatory signals that adjust neuronal gain, thereby altering neuronal responsiveness to incoming stimuli~\citep{ferguson_mechanisms_2020, shine_neuromodulatory_2019, aston-jones_integrative_2005}. For instance, during perceptual updating, the noradrenergic system originating from the LC transiently boosts neuronal gain to facilitate sensory integration of new incoming stimuli~\citep{wainstein_gain_2025, jordan_locus_2024, aston-jones_integrative_2005}. We propose that this neuronal gain modulation naturally gives rise to a fast-slow learning structure by shaping the effective synaptic weights. Consider a model $y(t) = h \left( x(t); g(t), w(t) \right)$  and a loss function $L \left( y(t), T(t) \right)$, where $x(t)$ represents the input at time $t$, $\{g(t), w(t)\}$ denotes the parameter estimate, $y(t)$ is the output produced by the model, and $T(t)$ is the target. In artificial neural networks, we can interpret $w_{ij} (t)$ as the synaptic weights connecting neuron $j$ with neuron $i$ and $g_i(t)$ as the neuronal gain; hence, one can consider an effective weight 
\begin{equation}
    W^{\text{eff}}_{ij}(t) = g_i(t)w_{ij}(t).
\end{equation}
Assuming neuronal gain follows noradrenergic activity, we distinguish a phasic mode of rapid bursts driven by prediction errors that transiently boost gain, and a tonic baseline mode $g_{0,i}$. This leads to a decomposition of the effective weight as contributions from these components, i.e. 
\begin{equation}
    W^{\text{eff}}_{ij}(t) = g_{i}(t)w_{ij}(t) = g_{0,i}w_{ij}(t) + \left[ g_i(t) - g_{0,i} \right] w_{ij}(t).
\end{equation}
This decomposition highlights how neuronal gain modulation inherently approximates a two-timescale optimization scheme. The slow component, $w_{ij}^{\text{slow}} (t) = g_{0,i}w_{ij}(t)$, represents the stable synaptic changes associated with tonic neuromodulation. Conversely, the fast component, $w_{ij}^{\text{fast}} (t) = \left[ g_i(t) - g_{0,i} \right] w_{ij}(t)$ dynamically emerges in response to phasic neuromodulatory signals, effectively isolating rapid contextual adaptations from stable memory traces. Simply, transient gain modulation ``virtually'' decouples the effective synaptic weight updates into slow and fast components, offering an adaptive mechanism to flexibly balance plasticity and stability (Figure~\ref{fig:multitimescale-weight}a).
To further illustrate this concept, following~\citet{jones_learning_2022}, we consider a simplified linear model defined by $y(t) = W_{\text{eff}} (t) \, x$ with a constant input $x \equiv 1$ to predict a target signal $T(t)$ under the mean squared error loss $L = \frac{1}{2} \left[ T(t) - y(t) \right]^2$. Parameter updates follow standard stochastic gradient descent, while gain dynamics evolve via exponential decay modulated by transient increases in response to contextual surprises \citep{wainstein_gain_2025} -- see Appendix~\ref{app:gain_dynamics} for a further discussion. Formally:
\begin{align}
    w(t+1) &= w(t) - \alpha\partial_w L \label{eq:weight_update}, \\
    g(t+1) &= \gamma g(t) + (1-\gamma) g_0 + \eta H(y)
    \label{eq:gain_update}
\end{align}
where $\alpha$ is the learning rate, $\gamma < 1$ controls how quickly the gain decays back to its tonic baseline~($g_0=1$ for simplicity), and $\eta$ determines the strength of the phasic bump, driven by the ``surprise'' signal $H(y)$, here approximated by the absolute value of the loss gradient $\lvert \partial_g L \rvert$, reflecting its link to error prediction~\citep{jordan_locus_2024}. Given that the gradient with respect to synaptic weights can be expressed as
\begin{equation}
\label{eq:base-scaling}
\frac{\partial L}{\partial w} = g(t)\frac{\partial L}{\partial W_\text{eff}},
\end{equation}
neuronal gain acts explicitly as a plasticity modulator, dynamically scaling the magnitude of learning updates. During high uncertainty or unexpected errors, phasic increases in neuronal gain amplify synaptic updates during backpropagation, allowing rapid adaptation to novel contexts. Moreover, these gain boosts induce a reparameterization of the loss~\citep{dinh_bengio_icml17}, which transiently flattens the loss landscape in the effective-weight space, as illustrated in Figure~\ref{fig:multitimescale-weight}b (right panel). Once confidence is restored, gain decays toward the baseline, promoting stability. Thus, gain modulation not only amplifies plasticity but also transiently flattens the loss surface, facilitating smoother task transitions.
\subsection{Gain rescales effective-weight geometries}
\label{sec:gain_reparameterization}

Beyond introducing fast and slow components of the effective synaptic weights, gain modulation establishes a distinction between the parameter space in which optimization is performed and the effective parameter space in which the network is evaluated. Gradient descent updates a set of base weights $w$, whereas the forward pass is computed using the effective weights $W_{\mathrm{eff}}=gw$. To formalize this geometric effect, let $\mathcal{W}\subseteq\mathbb{R}^n$ denote the base-weight space, let $\mathcal{W}_{\mathrm{eff}}\subseteq\mathbb{R}^n$ denote the effective-weight space, and let $L:\mathcal{W}_{\mathrm{eff}}\rightarrow\mathbb{R}$ be a loss function that is twice continuously differentiable in a local neighborhood of the point under consideration. We introduce the invertible linear reparameterization map
\begin{equation}
    \Phi:\mathcal{W}\rightarrow\mathcal{W}_{\mathrm{eff}},
    \qquad
    W_{\mathrm{eff}}=\Phi(w)=Gw,
    \label{eq:gain_reparam}
\end{equation}
where $G\in\mathbb{R}^{n\times n}$ is an invertible gain matrix. In the isotropic case considered below, $G=gI_n$, with scalar gain $g>0$. The loss expressed as a function of the base weights is then
\begin{equation}
    \widetilde{L}(w)=L\!\left(\Phi(w)\right)=L(Gw).
    \label{eq:reparameterized_loss}
\end{equation}
Here, $L(W_{\mathrm{eff}})$ denotes the loss expressed in the effective-weight coordinates used during the forward pass, whereas $\widetilde{L}(w)$ denotes the same objective expressed in the base-weight coordinates updated by gradient descent.

\paragraph{Lemma 1 (Gradient transformation).}
For every $w\in\mathcal{W}$,
\begin{equation}
    \nabla_w\widetilde{L}(w)
    =
    G^\top\nabla_{W_{\mathrm{eff}}}L(W_{\mathrm{eff}}).
    \label{eq:gradient_transformation}
\end{equation}
In the isotropic case $G=gI_n$,
\begin{equation}
    \nabla_w\widetilde{L}(w)
    =
    g\nabla_{W_{\mathrm{eff}}}L(W_{\mathrm{eff}}).
    \label{eq:isotropic_gradient}
\end{equation}
\emph{Proof.} This follows directly from the chain rule applied to Equation~\eqref{eq:reparameterized_loss}.

In the isotropic case, Lemma~1 recovers Equation~\eqref{eq:base-scaling}. The displacement induced by the base-weight update, expressed in effective-weight coordinates, is
\begin{equation}
    \Delta W_{\mathrm{eff}}^{\mathrm{weight}}=-\alpha g_t^2 \nabla_{W_{\mathrm{eff}}} L\!\left(W_{\mathrm{eff}}^{t}\right).
    \label{eq:effective_weight_gain_update}
\end{equation}
Hence, during the weight-update substep, gain induces an effective step size $\alpha_{\mathrm{eff}}(t)=\alpha g_t^2$ for the gradient-driven displacement in effective-weight coordinates. Positive scalar gain therefore enlarges the effective gradient step, which supports rapid adaptation under uncertainty.

\paragraph{Lemma 2 (Hessian transformation).}
For every $w\in\mathcal{W}$,
\begin{equation}
    H_w(w)
    =
    G^\top H_{\mathrm{eff}}(W_{\mathrm{eff}})G,
    \label{eq:hessian_transformation}
\end{equation}
where
\begin{equation}
    H_w(w)=\nabla_w^2\widetilde{L}(w),
    \qquad
    H_{\mathrm{eff}}(W_{\mathrm{eff}})
    =
    \nabla_{W_{\mathrm{eff}}}^2L(W_{\mathrm{eff}}).
    \label{eq:hessian_definitions}
\end{equation}
In the isotropic case $G=gI_n$,
\begin{equation}
    H_w(w)
    =
    g^2H_{\mathrm{eff}}(W_{\mathrm{eff}}).
    \label{eq:isotropic_hessian}
\end{equation}
\emph{Proof.} This follows from the second-order chain rule for the linear reparameterization $\Phi(w)=Gw$.

\paragraph{Corollary 1 (Eigenvalue rescaling).}
For any corresponding pair of parameters $w\in\mathcal{W}$ and $W_{\mathrm{eff}}=gw\in\mathcal{W}_{\mathrm{eff}}$, the eigenvalues of the two Hessians in the isotropic case satisfy
\begin{equation}
    \lambda_{\mathrm{eff},i}(W_{\mathrm{eff}})
    =
    \frac{1}{g^2}\lambda_{w,i}(w).
    \label{eq:eigenvalue_effective}
\end{equation}
Hence, relative to the local geometry expressed in the base-weight parameterization, the corresponding curvature is scaled by $g^{-2}$ in the effective-weight coordinates used during the forward pass. We refer to this coordinate-dependent curvature reduction as \emph{effective-coordinate flattening}.

Importantly, Equations~\eqref{eq:isotropic_hessian} and \eqref{eq:eigenvalue_effective} reflect the dependence of Hessian-based sharpness on parameterization~\citep{dinh_bengio_icml17}. NGM-SGD therefore combines an optimizer-side amplification of plasticity with effective-coordinate flattening in the parameterization in which the network is evaluated. Upon a change in the data distribution, we hypothesize that this coupled mechanism attenuates the transition-induced amplification of effective-coordinate curvature and previous-task loss that would otherwise accompany the enlarged effective step, without altering the instantaneous gradient direction.

\subsection{Noradrenergic uncertainty}
Biological and theoretical evidence converge on the necessity of explicitly representing a learner’s uncertainties when interacting with stochastic, non-stationary environments to achieve optimal inference~\citep{yu_uncertainty_2005, dayan_phasic_2006, jordan_locus_2024}. Noradrenaline has been particularly linked to ``unexpected uncertainty'', where its phasic signaling is triggered by sudden increases in environmental variability due to contextual changes~\citep{yu_uncertainty_2005, dayan_phasic_2006, mei_improving_2025, jordan_locus_2024}. This response injects additional variability into neural and behavioral processes, facilitating rapid adaptation to shifting contexts~\citep{wainstein_gain_2025, munn_ascending_2021, shine_neuromodulatory_2019}. Evidence in this regard from~\citet{wainstein_gain_2025} suggests that this form of uncertainty can be quantified as the Shannon entropy of the network’s readout:
\begin{equation}
    H(y) = - \sum_i \pi_i(y) \log\left( \pi_i(y) \right)
    \label{eq:uncertainty}
\end{equation}
where $\pi(y)$ denotes the softmax probability distribution of the network's output~\citep{shannon_mathematical_1948}. We incorporate this into Equation~\eqref{eq:gain_update} to dynamically modulate neuronal gain, which acts as a forcing function that transiently boosts gain in response to novel or ambiguous stimuli~\citep{jordan_locus_2024, aston-jones_integrative_2005, shine_neuromodulatory_2019}. To formalize this process, we present Algorithm~\ref{alg:ngm_sgd} and benchmark it against MSGD~\citep{rumelhart_learning_1986}, Adam~\citep{kingma_adam_2014}, and vanilla SGD~\citep{goodfellow_DL2016}. MSGD and Adam implicitly embody multi-timescale dynamics --  MSGD can be interpreted as a fast-slow separation where an exponentially decaying gradient trace smooths high-frequency noise~\citep{jones_learning_2022}, while Adam adapts step sizes based on local gradient statistics. Vanilla SGD is included as a non-adaptive single-timescale baseline.
\begin{algorithm}[!htbp]
\caption{\\ Noradrenergic Gain-Modulated SGD \textbf{(NGM‑SGD)}}
\label{alg:ngm_sgd}
\begin{algorithmic}[1]
\Require Dataset sequence $\mathcal{C} = \{c_k\}_{k=1}^{K}$, where each $c_k = \{(\mathbf{x}_j, \tilde{y}_j)\}_{j=1}^{N} \sim \mathcal{D}_k$; context iterations $I_{\mathcal{C}}$; batch size $B$; learning rate (lr) $\alpha > 0$; gain-decay $0 < \gamma < 1$; gain-baseline $g_{0} \geq 1$; entropy scale $\eta > 0$
\State \textbf{Initialize:} $\mathbf{W}\gets \mathbf{W}_{\mathrm{init}}$;\quad $g\gets g_{\mathrm{init}} = g_{0}$
\For{each context $c_k \in \mathcal{C}$}
    \For{iteration $i=1$ \textbf{to} $I_{\mathcal{C}}$}
        \State $(\mathbf{X},\tilde{\mathbf{Y}}) \sim \mathcal{D}_k^B$
        \Comment{\texttt{mini-batch}}
    
        \State $\boldsymbol{\pi} \gets \mathrm{softmax}\,\bigl(F_{\mathbf{W}}(\mathbf{X};\,g)\bigr)$        
        \Comment{\texttt{forward pass}}
    
        \State $\mathcal{L}\gets \tfrac{1}{B}\sum_{j=1}^{B}\ell(\boldsymbol{\pi}_j,\tilde{y}_j)$
        \Comment{\texttt{cross-entropy loss}}
    
        \State $\mathbf{W} \gets \mathbf{W} - \alpha\,\nabla_{\mathbf{W}}\mathcal{L}$
        \Comment{\texttt{SGD update}}
    
        \State $H\gets -\tfrac{1}{B}\sum_{j=1}^{B}\sum_{l}\,\pi_{j,l}\,\log\pi_{j,l}$
        \Comment{\texttt{uncertainty}}
    
        \State $g \gets \gamma\,g + (1-\gamma)\,g_{0} + \eta\,H$              
        \Comment{\texttt{gain update}}
    \EndFor
\EndFor
\end{algorithmic}
\end{algorithm}
\subsection{Mechanistic empirical validation}
\label{sec:mechanistic_validation}
\begin{figure*}[pos=!htbp]
  \centering
  \includegraphics[width=0.9\textwidth]{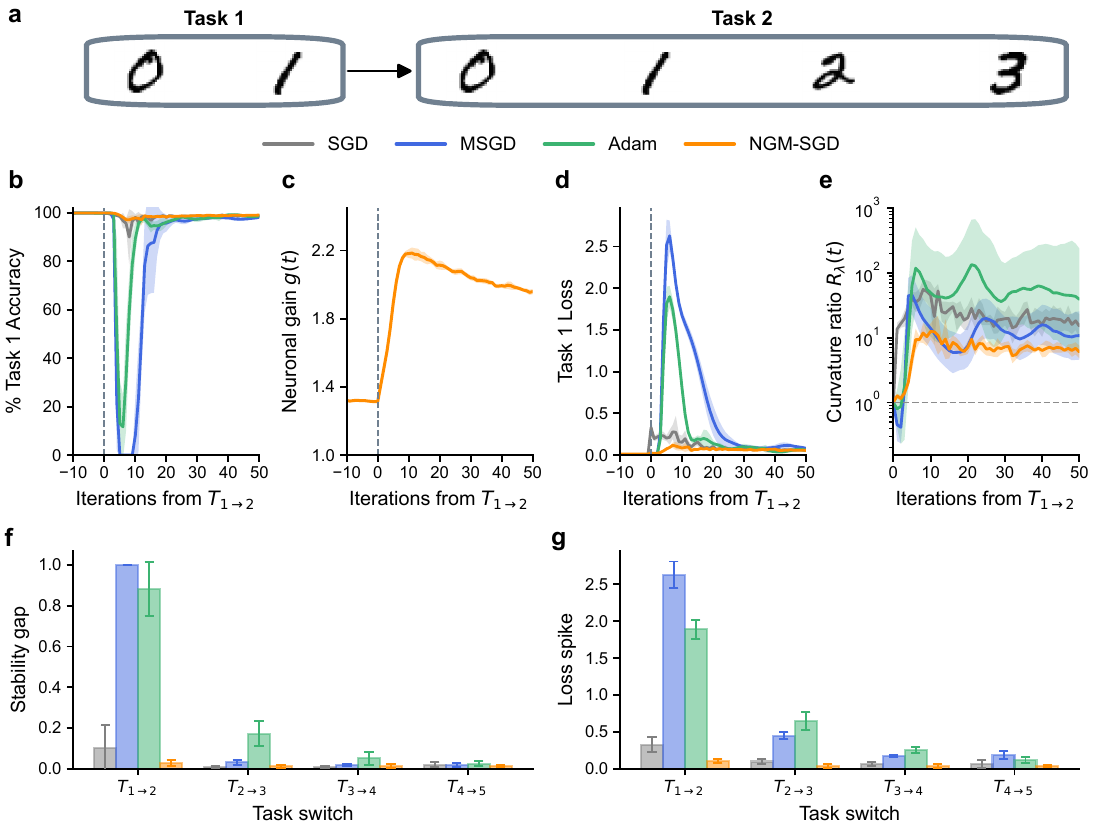}
    \caption{\textbf{Gain modulation attenuates transition-induced instability in Split MNIST.} \textbf{a)}~Samples of data around the first task transition, $T_{1\rightarrow2}$, where the cumulative training distribution expands from digits $\{0,1\}$ to $\{0,1,2,3\}$. \textbf{b)}~Test accuracy on Task~1 after the first task transition. \textbf{c)} Neuronal gain dynamics $g(t)$ for NGM-SGD, showing the phasic gain response following the task transition. \textbf{d)} Task~1 test loss increase after the task switch. \textbf{e)}~Effective-coordinate curvature ratio $R_{\lambda}(t)=|\lambda_{\mathrm{eff}}(t)|/|\lambda_{\mathrm{eff}}(0)|$, where $|\lambda_{\mathrm{eff}}(0)|$ denotes the effective curvature measured immediately before the transition.  The horizontal dashed line indicates $R_{\lambda}(t)=1$, corresponding to the pre-transition curvature. The vertical dashed lines in (b--d) mark the task switch at iteration zero. Curves show the mean, and shaded regions show the standard deviation across five independent runs. The curvature ratio is represented as the geometric mean $\pm 1$ standard deviation in log-space. \textbf{f--g)} Transition-level stability measures across the complete five-task sequence. \textbf{f)} Stability gap, computed as the maximum relative reduction in previous-task accuracy following each task switch (see Table~\ref{tab:stabgap_allmetrics}). \textbf{g)}~Previous-task loss spike, computed as the maximum post-transition loss minus the loss immediately before the switch. Bars and error bars denote the mean and standard deviation across five independent runs. Color coding denotes the different optimizers: our method (NGM-SGD, Algorithm~\ref{alg:ngm_sgd}) in orange, momentum-SGD (MSGD) in blue, Adam in green, and vanilla SGD in gray.}
  \label{fig:mechanistic_splitMNIST_T12}
\end{figure*}

Our hypothesis is rooted in gain modulation's ability to induce a reparameterization that increases the effective gradient-step magnitude while reducing the curvature in effective-weight coordinates. This dual action should lower test-set loss and interference at task transitions, thereby narrowing stability gaps during sequential training. Thus, to rigorously understand the impact of our approach in attenuating the stability gap, we isolate the stability gap from other sources of forgetting by using a joint-training experimental setting (Figure~\ref{fig:mechanistic_splitMNIST_T12}a); see Section~\ref{sec:methods} for further details of the experimental setting. We first use Split MNIST as a controlled proof-of-principle class-incremental benchmark, where the label space expands as new classes are introduced at each task~\citep{van_de_ven_three_2022}. This progressive incorporation of previously unseen categories provides a simple analogue of perceptual updating~\citep{wainstein_gain_2025,sales2019} and allows us to characterize the update-level dynamics of gain, previous-task performance, loss, and local curvature. We then perform targeted ablations to determine how gain reparameterization and its temporal dynamics contribute to the resulting stability-plasticity trade-off.
\subsubsection{Gain modulation reshapes optimization at task transitions}
We first examine the $T_{1\rightarrow2}$ transition in detail, where the cumulative training distribution expands from two to four classes (Figure~\ref{fig:mechanistic_splitMNIST_T12}a). This transition produces the largest relative change in the training distribution and the clearest separation between optimizers. Immediately after the task switch, MSGD and Adam exhibit pronounced transient reductions in Task~1 accuracy, whereas NGM-SGD largely preserves previous-task performance (Figure~\ref{fig:mechanistic_splitMNIST_T12}b). Vanilla SGD also exhibits substantially less degradation than the momentum-based optimizers, suggesting that the inertia terms of multi-timescale optimizers may introduce a further transition-induced instability (see Figure~\ref{fig:oracle_reset} and Appendix~\ref{sec:app_resetting}).

The stable response of NGM-SGD coincides with a phasic increase in neuronal gain following the transition (Figure~\ref{fig:mechanistic_splitMNIST_T12}c). As gain rises, NGM-SGD exhibits only a small increase in Task~1 test loss, in contrast to the large loss spikes produced by MSGD and Adam (Figure~\ref{fig:mechanistic_splitMNIST_T12}d). To examine whether this difference is associated with the geometric mechanism proposed in Section~\ref{sec:gain_reparameterization}, we measured the dominant-magnitude curvature of the Task~1 loss in effective-weight coordinates and normalized it by its pre-transition value within each run\footnote{Curvature was estimated using 25-step power iteration over five fixed Task~1 test batches.}. Although the distributional shift increases curvature for all optimizers, NGM-SGD produces a substantially smaller post-transition curvature amplification than MSGD and Adam and maintains a lower curvature ratio throughout the analyzed window (Figure~\ref{fig:mechanistic_splitMNIST_T12}e). This behavior is consistent with gain reparameterization producing a locally flatter loss landscape in effective-weight coordinates, thereby reducing the sensitivity of previous-task loss to the parameter displacements induced by the task transition.

We further examined whether the transition-level observations extended beyond the update-level dynamics of the first task switch. NGM-SGD exhibits consistently small Task~1 stability gaps and loss spikes across the complete task sequence (Figure~\ref{fig:mechanistic_splitMNIST_T12}f,g). However, the largest separation between optimizers occurs at $T_{1\rightarrow2}$, where MSGD and Adam show substantial transient instability while NGM-SGD remains close to the SGD baseline or improves upon it. The magnitude of these effects decreases at later transitions as each additional pair of classes constitutes a progressively smaller relative expansion of the cumulative training distribution. Consequently, averaging the stability gap across transitions can obscure the more pronounced reduction observed at the strongest distributional shift.

\subsubsection{Ablations show that NGM-SGD maintains the stability-plasticity balance}
\begin{figure*}[pos=!htbp]
  \centering
  \includegraphics[width=0.95\textwidth]{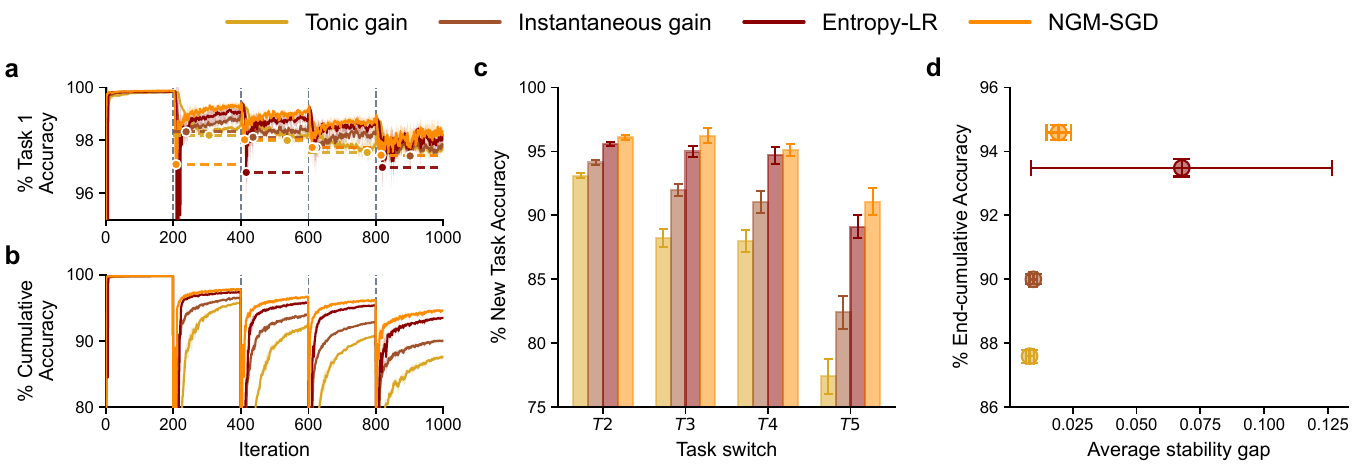}
  \caption{\textbf{Effect of ablating gain modulation on Split MNIST.} Comparison of the full NGM-SGD method (Algorithm~\ref{alg:ngm_sgd}, orange) with three targeted ablations: \textit{Tonic gain} (gold), which removes entropy-driven phasic modulation ($\eta = 0$); \textit{Instantaneous gain} (brown), which removes temporal integration of the gain dynamics ($\gamma = 0$); and \textit{Entropy-LR} (dark red), which applies the entropy-dependent signal directly to the learning rate ($\alpha(t)=\alpha_0 q(t)^2$). \textbf{a)} Test accuracy on Task~1 as the model is incrementally trained across all benchmark tasks. \textbf{b)} Cumulative test accuracy over all tasks observed up to the current training stage. \textbf{c)} End-of-task accuracy on the newly introduced task following each task switch. \textbf{d)} Trade-off between the average stability gap across task transitions and the final cumulative accuracy at the end of training. Curves represent the mean~$\pm$~standard deviation (shaded area) over five runs with different random seeds. Bars and error bars denote the corresponding mean~$\pm$~standard deviation. In panel (d), markers denote the mean values, while horizontal and vertical error bars indicate the propagated standard deviation of the average stability gap and the standard deviation of the final cumulative accuracy, respectively. Vertical dashed lines in panels (a--b) indicate task transitions. Horizontal dotted lines in panel a illustrate the min-ACC per task.}
  \label{fig:splitMNIST_gainablations}
\end{figure*}
To characterize how the dynamic gain reparameterization affects the stability-plasticity trade-off exhibited by our method, we compared NGM-SGD with three targeted ablations (Figure~\ref{fig:splitMNIST_gainablations}). \textit{Tonic gain} removes entropy-driven phasic modulation by setting $\eta=0$, \textit{Instantaneous gain} removes temporal integration by setting $\gamma=0$, and \textit{Entropy-LR} applies the entropy-dependent signal scaling directly to the learning rate, i.e. $\alpha(t)=\alpha_0 q(t)^2$ -- where $q(t)$ follows the same dynamics as Equation~\eqref{eq:gain_update} -- matching the gradient-driven displacement induced by NGM-SGD in effective-weight coordinates.

Tonic and instantaneous gain preserve high Task~1 accuracy and produce comparatively small transient drops across task transitions (Figure~\ref{fig:splitMNIST_gainablations}a). However, this apparent increase in stability results from weaker acquisition of the newly introduced tasks, leading to lower cumulative and new-task accuracies, as shown in Figure~\ref{fig:splitMNIST_gainablations}b,c. For tonic gain, the absence of dynamic gain updates eliminates the phasic component of the effective weight, leaving only the slow contribution, $w^{\mathrm{slow}}_{ij}=g_{0,i}w_{ij}$. Without a fast adaptation channel, newly introduced classes are learned more slowly, creating the appearance of improved stability. Similarly, removing temporal integration in the instantaneous-gain condition prevents the gain response from being sustained across successive updates, limiting its capacity to support rapid acquisition. Their small stability gaps therefore reflect reduced plasticity rather than an improved stability-plasticity trade-off.

Entropy-LR exhibits the opposite behavior. Applying the entropy signal directly to the learning rate supports stronger acquisition of new tasks, but results in the largest average stability gap (Figure~\ref{fig:splitMNIST_gainablations}c,d). This resembles the behavior observed with MSGD, as both methods introduce temporally integrated, two-timescale adaptation of the effective update magnitude~\citep{jones_learning_2022}, which can accelerate learning but also amplify transient instability at task transitions. Entropy-dependent step-size adaptation alone is therefore insufficient to reproduce the transition-stabilizing effect of gain modulation. Although both mechanisms alter the magnitude of the parameter updates, neuronal gain additionally reparameterizes the network during the forward pass and modifies the optimization geometry in the effective-weight coordinates.

NGM-SGD achieves the most favorable stability-plasticity trade-off, combining a low average stability gap with the highest final cumulative accuracy, as shown most clearly in Figure~\ref{fig:splitMNIST_gainablations}d. In contrast, tonic and instantaneous gain favor stability at the expense of cumulative acquisition, whereas Entropy-LR improves plasticity but produces substantially greater transition-induced instability. This contrast highlights the importance of the complete dynamic gain mechanism: phasic gain modulation engages a fast adaptation component that facilitates the integration of new information, while its temporal dynamics regulate this response across successive updates. Together with gain-induced reparameterization and flattening of the effective loss landscape, these properties jointly support rapid adaptation while limiting transient degradation of previously acquired performance.

\subsection{Online task-agnostic continual learning benchmarks}
\begin{table*}[t]
  \centering
  \caption{\textbf{Main quantitative metrics.} For all benchmarks we report (across tasks) the average final accuracy (avg-ACC), average minimum accuracy (avg-min-ACC), average stability gap drop (avg-SG), and the worst-case accuracy (WC-ACC). Highlighted values indicate the best results.}  
  \label{tab:ce_metrics}
  \centering
  \resizebox{\textwidth}{!}{
  \begin{tabular}{@{}cccccc@{}}
    \toprule
     &  & \textbf{avg-ACC ($\uparrow$)} & \textbf{avg-min-ACC ($\uparrow$)} & \textbf{WC-ACC ($\uparrow$)} & \textbf{avg-SG ($\downarrow$)} \\
    \midrule 
    \multirow[c]{4}{*}{\shortstack{Split\\MNIST}}
      & NGM-SGD (ours) & 99.015 $\pm$ 0.047 & \textbf{97.570 $\pm$ 0.465} & \textbf{97.696 $\pm$ 0.373} & \textbf{0.017 $\pm$ 0.005} \\
      & MSGD & 98.971 $\pm$ 0.086 & 72.518 $\pm$ 0.403 & 77.628 $\pm$ 0.329 & 0.267 $\pm$ 0.004 \\
      & Adam & \textbf{99.088 $\pm$ 0.108} & 71.047 $\pm$ 3.753 & 76.537 $\pm$ 3.003 & 0.283 $\pm$ 0.038 \\
      & SGD & 98.975 $\pm$ 0.099 & 95.887 $\pm$ 2.887 & 96.293 $\pm$ 2.311 & 0.034 $\pm$ 0.029 \\
    \midrule
    \multirow[c]{4}{*}{\shortstack{Split\\CIFAR-10}}
      & NGM-SGD (ours) & \textbf{90.630 $\pm$ 1.576} & \textbf{79.485 $\pm$ 3.875} & \textbf{80.880 $\pm$ 3.349} & \textbf{0.134 $\pm$ 0.043} \\
      & MSGD & 89.434 $\pm$ 2.134 & 52.828 $\pm$ 3.898 & 58.348 $\pm$ 3.521 & 0.425 $\pm$ 0.044 \\
      & Adam & 90.342 $\pm$ 1.579 & 63.450 $\pm$ 7.839 & 67.692 $\pm$ 6.335 & 0.310 $\pm$ 0.088 \\
      & SGD & 89.478 $\pm$ 1.909 & 68.712 $\pm$ 4.958 & 71.868 $\pm$ 4.007 & 0.241 $\pm$ 0.059 \\
    \midrule
    \multirow[c]{4}{*}{\shortstack{Split\\mini-ImageNet}}
      & NGM-SGD (ours) & \textbf{48.312 $\pm$ 2.289} & \textbf{33.165 $\pm$ 2.435} & \textbf{36.040 $\pm$ 2.058} & \textbf{0.300 $\pm$ 0.071} \\
      & MSGD & 46.824 $\pm$ 2.176 & 26.775 $\pm$ 2.417 & 30.416 $\pm$ 2.029 & 0.409 $\pm$ 0.067 \\
      & Adam & 47.208 $\pm$ 2.044 & 28.300 $\pm$ 3.371 & 32.052 $\pm$ 2.768 & 0.384 $\pm$ 0.077 \\
      & SGD & 42.988 $\pm$ 2.549 & 28.345 $\pm$ 3.671 & 30.288 $\pm$ 3.309 & 0.332 $\pm$ 0.098 \\
    \midrule
    \multirow[c]{4}{*}{\shortstack{Rotated\\MNIST}}
      & NGM-SGD (ours) & 94.948 $\pm$ 0.121 & \textbf{90.015 $\pm$ 1.451} & \textbf{91.542 $\pm$ 0.973} & 0.054 $\pm$ 0.015 \\
      & MSGD & 95.297 $\pm$ 0.217 & 84.329 $\pm$ 1.359 & 87.823 $\pm$ 0.914 & 0.117 $\pm$ 0.014 \\
      & Adam & \textbf{95.777 $\pm$ 0.195} & 87.177 $\pm$ 1.060 & 89.910 $\pm$ 0.721 & 0.092 $\pm$ 0.011 \\
      & SGD & 94.139 $\pm$ 0.255 & 89.107 $\pm$ 1.935 & 90.787 $\pm$ 1.300 & \textbf{0.053 $\pm$ 0.021} \\
    \midrule
    \multirow[c]{4}{*}{\shortstack{Domain\\CIFAR-100}}
      & NGM-SGD (ours) & 68.700 $\pm$ 0.881 & \textbf{53.420 $\pm$ 2.804} & \textbf{58.290 $\pm$ 1.916} & \textbf{0.224 $\pm$ 0.042} \\
      & MSGD & 67.160 $\pm$ 0.994 & 51.370 $\pm$ 2.234 & 56.867 $\pm$ 1.556 & 0.232 $\pm$ 0.037 \\
      & Adam & \textbf{69.163 $\pm$ 0.871} & 51.615 $\pm$ 2.653 & 57.323 $\pm$ 1.840 & 0.256 $\pm$ 0.040 \\
      & SGD & 65.463 $\pm$ 1.204 & 49.050 $\pm$ 3.177 & 54.950 $\pm$ 2.229 & 0.243 $\pm$ 0.052 \\
    \bottomrule
  \end{tabular}
  }
\end{table*}
We next assess whether the effects observed on Split MNIST generalize to other class-incremental benchmarks, namely Split CIFAR-10 and Split mini-ImageNet. Additionally, we consider a complementary continual-learning scenario: domain-incremental learning~\citep{van_de_ven_three_2022}. For this setting, we evaluate Rotated MNIST and Domain CIFAR-100, where the label space remains fixed while the input distribution changes across tasks. NGM-SGD is compared with tuned SGD, MSGD, and Adam baselines in terms of transient stability and final predictive performance.

In Table~\ref{tab:ce_metrics}, we summarize the main experimental results across benchmarks using standard continual evaluation metrics established early in the field \citep{de_lange_continual_2022} -- see Appendix~\ref{sec:metrics}. The final average accuracy (avg-ACC) reflects the model’s overall performance by averaging classification accuracy across all tasks at the end of training~\citep{hess_two_2023}. The average minimum accuracy (avg-min-ACC) captures the lowest accuracy reached per task, averaged across tasks~\citep{hess_two_2023}, and serves as a measure of transient forgetting. The worst-case accuracy (WC-ACC) incorporates the minimum accuracy on prior tasks and the accuracy on the current task, offering a measure of the stability-plasticity trade-off~\citep{lapacz_exploring_2024, de_lange_continual_2022}. Lastly, the stability gap (avg-SG) represents the average maximum drop in accuracy observed when learning subsequent tasks~\citep{lapacz_exploring_2024, harun_overcoming_2023}, highlighting how much the model degrades when acquiring new tasks.
\begin{figure*}[pos=!htbp]
  \centering
  \includegraphics[width=0.95\textwidth]{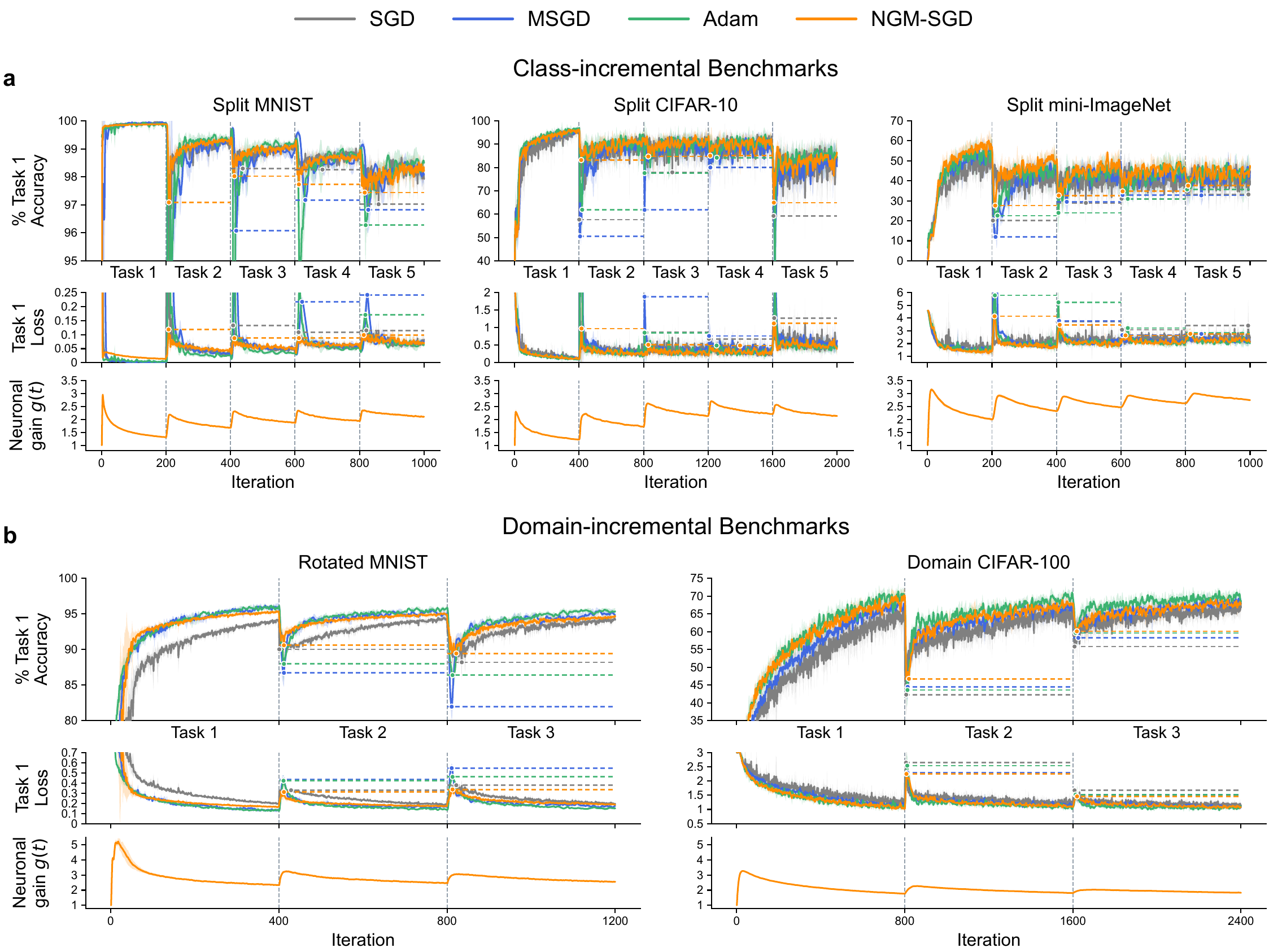}
    \caption{\textbf{Stability gaps across class- and domain-incremental learning.} \textbf{a)} Class-incremental results on Split MNIST, Split CIFAR-10, and Split mini-ImageNet. \textbf{b)} Domain-incremental results on Rotated MNIST and Domain CIFAR-100. Within each block, the top row shows Task~1 test accuracy as the model is trained incrementally across all benchmark tasks, the middle row shows the corresponding Task~1 test loss, and the bottom row shows the evolution of neuronal gain. Curves and shaded regions represent the mean~$\pm$~standard deviation over five runs with different random seeds, and dots indicate the minimum Task~1 accuracy reached within each task. NGM-SGD is shown in orange, momentum-SGD in blue, Adam in green, and vanilla SGD in gray. Accuracy panels are vertically restricted to emphasize the stability gaps; consequently, some performance drops may appear truncated.}
  \label{fig:incremental_benchmarks}
\end{figure*}
\subsection{NGM-SGD reduces stability gaps}
\begin{figure*}[pos=!htbp]
  \centering
  \includegraphics[width=\textwidth]{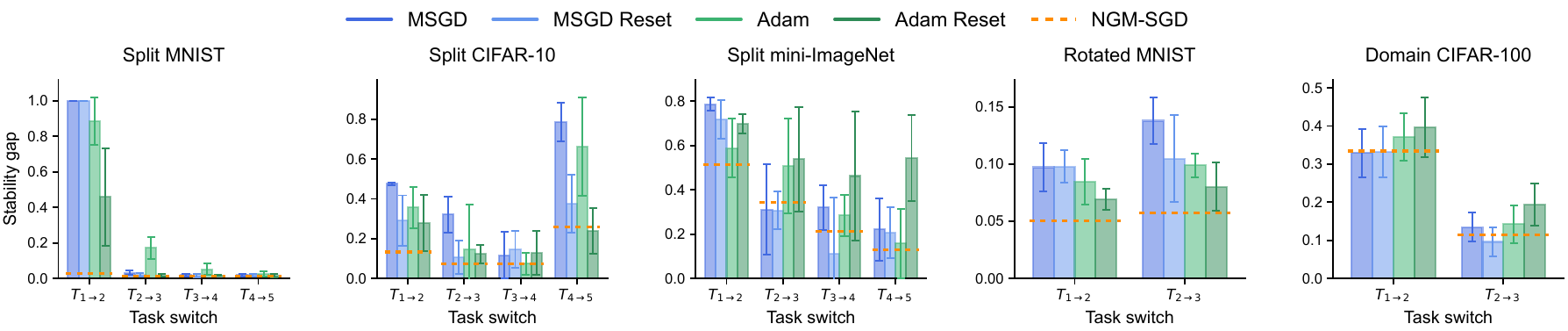}
    \caption{\textbf{Effect of resetting optimizer inertia at task transitions.} Stability gaps for MSGD and Adam, with and without resetting their accumulated optimizer states at each task switch, across Split MNIST, Split CIFAR-10, Split mini-ImageNet, Rotated MNIST, and Domain CIFAR-100. Bars and error bars represent the mean~$\pm$~standard deviation over five runs with different random seeds. Orange dotted segments indicate the corresponding NGM-SGD stability gap at each transition.}
  \label{fig:oracle_reset}
\end{figure*}

Table~\ref{tab:ce_metrics} shows that NGM-SGD consistently reduces the stability gap, although the performance gains are less pronounced than hypothesized in prior work~\citep{hess_two_2023}. In class-incremental tasks (Figure~\ref{fig:incremental_benchmarks}a, top), the gap is largest at the $T_1 \rightarrow T_2$ transition, where the data distribution doubles in complexity (see Table~\ref{tab:stabgap_allmetrics} in Appendix~\ref{sec:metrics}); it then diminishes as later tasks alter the loss landscape less drastically. In Split CIFAR-10, forgetting is minimal in tasks 2-4 (animal classes distinct from the initial vehicle task\footnote{Task 1 consists of two vehicle categories from CIFAR-10.}), but reappears when vehicles return in the final task. By contrast, mini-ImageNet classes span diverse domains, masking this domain-specific effect. Domain-incremental tasks (Figure~\ref{fig:incremental_benchmarks}b, top) always exhibit interference due to overlapping feature spaces. Across both settings, stability gaps are most evident with multi-timescale optimizers such as Adam and MSGD, where momentum-driven updates overshoot at task switches, and abrupt loss-landscape shifts trigger transient forgetting. Vanilla SGD stays closer to the true gradient and generally shows smaller gaps (except in high-noise regimes). NGM-SGD further reduces drops by flattening the effective loss landscape on $T_1$, enabling stable evaluation across tasks while still supporting rapid acquisition of new knowledge. Additional analyses in Appendix~\ref{sec:app_resetting} show that resetting the inertia terms of Adam and MSGD at task transitions frequently attenuates the overshooting and reduces stability gaps, although both still underperform our NGM-SGD method (Figure~\ref{fig:oracle_reset}). Notably, in CIFAR and mini-ImageNet, gain modulation was applied only to the ResNet-18 classification head and still substantially mitigated stability gaps. This is consistent with the primary finding of \citet{lapacz_exploring_2024}, who noted that a sizable component of the stability gap can be attributed to the final classification layer.
\subsection{Gain boosts reduce test loss at task boundaries}
Aiming to understand the origin of NGM-SGD’s ability to reduce stability gaps, we examined the evolution of the $T_1$ loss across all experiments. As shown in Figures~\ref{fig:incremental_benchmarks}a and \ref{fig:incremental_benchmarks}b, NGM-SGD consistently lowers the test loss at task transitions compared to Adam and MSGD. We hypothesize that this effect arises from the gain reparameterization applied to the forward pass, which flattens the effective landscape of the test-set loss~\citep{dinh_bengio_icml17}. Notably, the learning trajectory followed by NGM-SGD may not correspond to the non-increasing loss path suggested in recent studies~\citep{hess_two_2023, kamath_expanding_2024}. However, unlike Adam and MSGD -- whose momentum terms alter the update direction -- NGM-SGD follows the true gradient, as the gain only rescales the step magnitude without deviating from the gradient itself. This, combined with the locally flattened Hessian induced by the gain reparameterization, allows the model to incorporate new information more smoothly and avoid sharp increases in loss at task switches.

\subsection{Neuronal gain reflects task-related complexity} 
In NGM-SGD, neuronal gain evolves according to Equations~\eqref{eq:gain_update} and \eqref{eq:uncertainty}, rising in response to task-related uncertainty and decaying as learning progresses~\citep{wainstein_gain_2025}. As shown in the bottom row of Figures~\ref{fig:incremental_benchmarks}a and \ref{fig:incremental_benchmarks}b, the asymptotic level to which gain decays differs between class-incremental and domain-incremental settings, reflecting distinct patterns of task complexity. In class-incremental scenarios, gain rises at the beginning of each new task and then decays to a progressively higher baseline. This indicates that the system perceives each new task as increasingly complex, requiring higher sustained sensitivity. In contrast, in domain-incremental tasks, gain also spikes at the onset of each task, but consistently decays to a similar level across tasks. This suggests that most of the complexity is already captured in the first task, and the subsequent tasks do not introduce additional uncertainty. 

\begin{figure}[pos=!htbp]
  \centering
  \includegraphics[width=0.9\columnwidth]{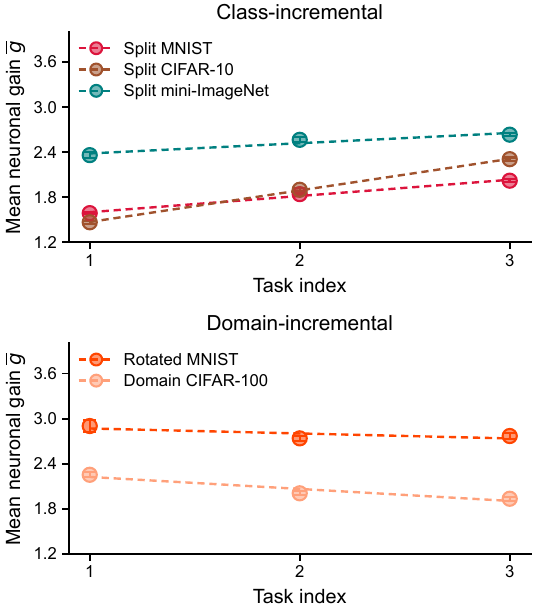}
  \caption{\textbf{Gain across tasks.} Plots of the mean neuronal gain in the first three tasks for class-incremental (top) and domain-incremental (bottom) experiments. We used a linear fit and obtained the slopes $2.16 \times 10^{-1}$ for Split MNIST, $4.20 \times 10^{-1}$ for Split CIFAR-10, $1.36 \times 10^{-1}$ for Split mini-ImageNet, $-6.63 \times 10^{-2}$ for Rotated MNIST, and $-1.59 \times 10^{-1}$ for Domain CIFAR-100.}
  \label{fig:gtaskComplexity}
\end{figure}

To quantify these differences, we computed the mean gain during each of the first three tasks and fitted a linear trend across tasks (Figure~\ref{fig:gtaskComplexity}). We focus on the first three tasks since, in the class-incremental experiments, the relative novelty introduced by Tasks 4 and 5 is smaller, and the changes in sustained gain are therefore less pronounced. Split MNIST, Split CIFAR-10, and Split mini-ImageNet exhibit positive slopes, indicating that sustained gain increases as additional classes and task demands accumulate. By contrast, Rotated MNIST and Domain CIFAR-100 exhibit slightly negative slopes, consistent with the absence of a progressive increase in complexity across domains.
Together, these results suggest that the long-term gain level can serve as a proxy for accumulated task uncertainty and complexity: it increases when successive tasks expand the classification problem, but remains comparatively stable when tasks preserve the same underlying class structure and primarily alter the input domain. This interpretation echoes evidence that biological systems regulate internal gain in response to changing cognitive demands~\citep{aston-jones_integrative_2005,ferguson_mechanisms_2020}.

\section{Discussion}
Our work contributes to the recent CL paradigm shift in perspective from \textit{what} to optimize in CL to \textit{how} to optimize it~\citep{hess_two_2023}. Inspired by noradrenergic phasic bursts that enhance cortical responsiveness during uncertainty~\citep{jordan_locus_2024, shine_neuromodulatory_2019}, we propose gain modulation as a biologically inspired mechanism that reframes optimization.  Rather than adapting through a fixed parameterization, uncertainty-driven gain boosts dynamically reshape the learning dynamics, supporting rapid adaptation while limiting transition-induced interference. Our mechanism has two key effects: \textit{(i)}~transient gain boosts create emergent fast and slow components of the effective synaptic weights~\citep{hinton_using_1987}, also increasing the effective magnitude of gradient-driven adaptation; and \textit{(ii)}~they induce a forward-pass reparameterization in which increases in gain at task switches produce effective-coordinate flattening~\citep{dinh_bengio_icml17}, and empirically reduce test loss and interference in a manner consistent with theoretical neuroscience models~\citep{wainstein_gain_2025, munn_ascending_2021}. While biological gain modulation is implemented through changes in spiking activity~\citep{munn_neuronal_2023, whyte_burst-dependent_2023, rodriguezgarcia2025rolegainneuromodulationlayer5}, we show that the same principle can be instantiated in standard multilayer perceptrons (MLPs) by scaling each neuron's incoming weights with a gain coefficient, providing a minimal implementation of gain modulation in conventional FFNN architectures.

The same formulation also suggests a specific extension to recurrent neural networks (RNNs), in which gain modulation could act simultaneously on neural-state and synaptic dynamics. In the pretrained RNN of \citet{wainstein_gain_2025}, the transition structure is already embedded in the supervised task, such that uncertainty-driven gain primarily destabilizes the attractor supporting the currently dominant percept and accelerates the transition into the alternative recurrent regime. Interestingly, \citet{Capone26} broadened this role of gain by showing that gain-modulated RNNs can support in-context learning and online adaptation across temporal and reinforcement-learning (RL) tasks while keeping their synaptic parameters fixed. However, this activity-based account leaves open how rapid gain-mediated adaptation interacts with persistent synaptic learning. In biological circuits, these processes are unlikely to be independent. Gain-dependent changes in neuronal excitability can regulate the rate of activity-dependent synaptic modification~\citep{rodriguezgarcia2025rolegainneuromodulationlayer5}, while response modulation can shape the activity correlations that guide otherwise unsupervised Hebbian learning~\citep{swinehart_supervised_2005}. Together, these observations suggest that transient modulation may both reconfigure ongoing computation and determine how that computation becomes progressively embedded in the synaptic weights. NGM-SGD provides an abstraction of this coupling, whereby the same uncertainty-driven gain signal immediately reconfigures the effective network while simultaneously regulating the persistent weight changes through which new information is consolidated.

Extending this mechanism to RL-based computational neuroscience settings would test whether gain-mediated transitions between recurrent regimes can be consolidated through synaptic learning without disrupting previously acquired dynamics. In this setting, the stability gap could be viewed as a failure to coordinate fast network reconfiguration with slower synaptic adaptation. NeuroGym~\citep{crocioni_neurogym_2026} provides a standardized framework for testing this idea, particularly in reversal-learning tasks, where contingency changes produce transient performance impairments followed by adaptation~\citep{kiyama_simple_2025}. Gain-modulated RNNs could test whether uncertainty-linked gain transients predict faster state reconfiguration, accelerated synaptic updating, and reduced post-switch impairment, generating hypotheses for noradrenergic recordings or manipulations in analogous animal paradigms. Nevertheless, it is important to note that performance in animal reversal-learning paradigms is measured under the currently rewarded contingency and is therefore more directly comparable to post-switch adaptation than to the previous-task accuracy used to quantify stability gaps in CL. Isolating retention of the preceding rule would require dedicated probe, reinstatement, or context-cued trials.

From a broader ML perspective, our results indicate that uncertainty can regulate not only the magnitude of a parameter update, but also the parameterization through which that update is expressed. This distinction is supported by our ablations (Figure~\ref{fig:splitMNIST_gainablations}): applying the entropy signal directly to the learning rate increases plasticity but does not reproduce the transition-stabilizing effect of neuronal gain, whereas gain modulation jointly alters neuronal sensitivity, effective-weight dynamics, and local optimization geometry. These findings suggest a broader design principle for adaptive learning systems, in which uncertainty controls both the timescale and the parameterization of learning rather than acting solely as a step-size signal. Because NGM-SGD operates at the optimizer level, it could be combined with methods targeting complementary sources of CL failure, including experience replay approaches~\citep{rolnick2019experiencereplaycontinuallearning}. More broadly, gain modulation may be useful whenever rapid adaptation must occur without destabilizing previously acquired competence, including continual pretraining~\citep{guo_efficient_2024} and few-shot adaptation~\citep{vinyals2016matching}. In such settings, the contribution of NGM-SGD should be understood primarily as improving robustness along the adaptation trajectory, rather than increasing asymptotic accuracy.

Interestingly, prior studies have shown that, in settings where stability gaps occur, a monotonically decreasing path through the loss landscape nevertheless exists, suggesting that these gaps are in principle avoidable \citep{hess_two_2023, kamath_expanding_2024}. Our approach does not require constraining the optimization trajectory to such a path, as gain modulation reshapes the geometry of the effective loss landscape itself, providing a complementary route to stability. Moreover, although our approach does not yield the improvement in final performance hypothesized as a potential benefit of reducing the stability gap \citep[H3;][]{hess_two_2023}, the attenuation of stability gaps it provides remains crucial in real-world continual learning, where inference frequently occurs before training has converged and transient instabilities directly affect reliability.
\paragraph{Limitations and future work.} We acknowledge that our experiments rely on MNIST and CIFAR and evaluate performance over a relatively small number of tasks, with mini-ImageNet included as a more challenging benchmark, but likewise assessed over a few tasks. However, this design is intentional. Using a small task set ensures that each task induces a substantial distributional shift, making the stability gap large enough to measure reliably, whereas adding many tasks would diminish these shifts and obscure the very phenomenon we aim to study. Furthermore, the joint-training setup isolates the transient, optimization-driven forgetting that defines the stability gap, preventing confounds from other forms of interference present in more realistic CL settings. This controlled setting also suggests how NGM-SGD's ability to mitigate the stability gap may scale to more complex datasets -- the entropy-driven gain scale ($\eta$) at task boundaries can be tuned according to the complexity of the dataset considered (see Figure~\ref{fig:app_gainEMA_H}a in Appendix~\ref{app:gain_dynamics}).

Evaluating NGM-SGD in more realistic CL settings would be valuable, for example, by incorporating partial replay. Importantly, the additional experiments included in Appendix~\ref{app:replay} (Table~\ref{tab:ER_metrics}) show that our method continues to attenuate the stability gap even when replay is used, confirming that the benefits of the gain dynamics persist beyond the idealized joint training setup. However, a particularly promising direction is to explore how cholinergic neuromodulation, which is central for attention and flexible learning, could support task segregation \citep{shine_neuromodulatory_2019}. This could be implemented through gain-gating mechanisms that temporarily silence subsets of neurons and block their updates in a sparse setting, helping to regularize learning and reduce catastrophic forgetting. Combined with transient gain boosts, such mechanisms could guide gradient updates more effectively and enable more robust \emph{online} CL.

\section{Methods} \label{sec:methods}
We consider a joint-training scenario in which a learner $F_{\theta}$ is trained on a sequence of $K$ contexts $\mathcal{C} = \{c_k\}_{k=1}^K$. Each context $c_k$ is a disjoint dataset $\{(\mathbf{x}_j, \tilde{y}_j)\}_{j=1}^N \sim \mathcal{D}_k$. Contexts are presented one after another in an online fashion, with each context trained for a fixed number of iterations $I_{\mathcal{C}}$ \footnote{Due to random sampling, some training images may be encountered more than once, although data augmentation (crops and flips) ensures that repeated samples may differ across occurrences.}. Since this is a joint-training setup, the model retains access to all data from previous contexts throughout learning.
\paragraph{Datasets.} Class-incremental learning experiments used the MNIST \citep{lecun_gradient-based_1998}, CIFAR-10 \citep{krizhevsky_learning_2009}, and mini-ImageNet  \citep{vinyals2016matching, russakovsky2015imagenet} datasets. Split MNIST and Split CIFAR-10 were constructed by dividing each dataset into five sequential tasks, each comprising a disjoint subset of the 10 target classes~\citep{hess_two_2023}. For Split mini-ImageNet, the 100 classes were heterogeneously divided into ten tasks of ten classes each, from which we selected five contexts \citep{hess_two_2023}. For domain-incremental learning, we considered MNIST~\citep{lecun_gradient-based_1998} and CIFAR-100~\citep{krizhevsky_learning_2009}. We defined Rotated MNIST with three tasks, each using the full MNIST dataset rotated by a fixed angle: ${0^\circ, 80^\circ, 160^\circ}$. These angles were chosen to maximize stability gaps while avoiding digit ambiguities, such as confusing 6 and 9 when rotated $180^\circ$~\citep{de_lange_continual_2022, hess_two_2023}. We define Domain CIFAR-100 by organizing tasks according to the coarse semantic categories in CIFAR-100~\citep{hess_two_2023}. Specifically, each of the twenty CIFAR-100 super-classes is split across three tasks, with each task containing one class from every super-class (20 classes total) to predict the correct super-class.
\paragraph{Mechanistic analyses.} For the Split-MNIST mechanistic experiments, we compared NGM-SGD with three targeted ablations: \emph{Tonic gain} ($\eta=0$), \emph{Instantaneous gain} ($\gamma=0$), and \emph{Entropy-LR}, which applies the uncertainty-dependent signal directly to the learning rate as $\alpha(t)=\alpha_0 q(t)^2$, with $q(t)$ following the gain dynamics in Equation~(4). For the $T_{1\rightarrow2}$ transition, we additionally estimated the dominant-magnitude Hessian eigenvalue of the Task~1 loss in effective-weight coordinates after each of the first 50 updates. For NGM-SGD, the instantaneous gains were absorbed into the weights, $W_{\mathrm{eff}}=\mathrm{diag}(g)W$, and fixed to one; curvature was estimated using 25 power-iteration steps over five fixed Task~1 test batches and normalized within each run as $R_\lambda(t)=(|\lambda_{\mathrm{eff}}(t)|+\epsilon)/(|\lambda_{\mathrm{eff}}(0)|+\epsilon)$.
\paragraph{Continual learning setup.} We employed continual evaluation with $\rho_{\text{eval}}=1$ and tested on the full test set of the first task~\citep{de_lange_continual_2022}. Batch sizes were chosen as 128 for experiments on the MNIST dataset and 256 for experiments on the CIFAR-10, CIFAR-100, and mini-ImageNet datasets. Per experiment, the number of iterations per context ($I_{\mathcal{C}}$) was selected as 200 for Split MNIST and Split mini-ImageNet, 400 for Rotated MNIST and Split CIFAR-10, and 800 for Domain CIFAR-100. Each experiment was repeated five times with different random seeds. For each metric, we report the mean and the standard deviation across runs.
\paragraph{Architectures.} For tasks based on the MNIST dataset, we used a feedforward neural network (FFNN) with two hidden layers of 400 units each~\citep{de_lange_continual_2022}. All neurons in the hidden and output layers were gain-modulated, where gain represents the slope of the neuron's input-output response curve. We employed the ReLU activation function and removed all biases. For CIFAR-10, CIFAR-100, and mini-ImageNet datasets, the increased visual complexity motivated the use of convolutional layers, so we adopted a Slim ResNet-18 backbone~\citep{lopez-paz_gradient_2017, de_lange_continual_2022}. Convolutional networks mirror the brain’s hierarchy of feature detectors, but unlike biological neurons, where each has its own input-output curve, CNN feature maps use a shared filter, so uniform scaling cannot match sensitivity at the neuron level and may disrupt useful feature learning (see Appendix~\ref{app:layer_freezing}). Thus, we restricted gain modulation to the output layer\footnote{In the NGM‑SGD cases on Slim ResNet-18, we trained the backbone with MSGD to maintain the network's stability in learning features with the same learning rate.}, where it could regulate input sensitivity without disrupting feature selectivity, consistent with biological neurons~\citep{ferguson_mechanisms_2020}. This choice is also supported by findings that link the classification layer to the stability gap phenomenon~\citep{lapacz_exploring_2024}.
\paragraph{Optimization.} NGM-SGD optimization followed Algorithm~\ref{alg:ngm_sgd}. To determine its hyperparameters, we conducted a soft sweep over the gain-decay parameter $\gamma \in \{0.8, 0.9, 0.95\}$ and the gain scale $\eta \in [0.1, 0.5]$ (see Appendix~\ref{sec:hyperparams}). The choice of $\gamma=0.9$ ensured that the transient gain decayed slowly enough, allowing the phasic gain response to persist across successive updates, while remaining within a bounded range to support long-term convergence. Dataset-specific $\eta$ values were set to $\{0.5, 0.4, 0.2, 0.1, 0.1\}$ for Rotated MNIST, Split MNIST, Split CIFAR-10, Domain CIFAR-100, and Split mini-ImageNet, respectively. Restricting $\eta$ to this range kept the instantaneous gain $g(t)=g_0+\Delta g$ low, which prevented unstable learning dynamics -- a constraint that can be grounded in biological settings, as cortical gain is bounded by spiking saturation~\citep{munn_neuronal_2023} -- while allowing us to tune it according to the dataset complexity to generate sizable gain transients that reduce the stability gap (see Figure~\ref{fig:app_gainEMA_H}a in Appendix~\ref{app:gain_dynamics}). For comparison, we considered MSGD~\citep{rumelhart_learning_1986} (momentum of $0.9$), vanilla SGD~\citep{goodfellow_DL2016} and Adam~\citep{kingma_adam_2014} (betas of $0.9,0.99$) optimizers. We selected the learning rate of each optimizer by sweeping $\text{lr}=\{1.0, 0.1, 0.01, 0.001, 0.0001\}$ and found that the best values were $0.1$ for vanilla SGD, $0.01$ for NGM-SGD and MSGD ($0.1$ in Rotated MNIST), and $0.001$ for Adam (see Appendix~\ref{sec:hyperparams}). All other training details were kept identical across optimizers to ensure fair comparison.  
%

\printcredits

\section*{Acknowledgments}
This work was supported by a Marie Skłodowska‐Curie Global Fellowship Agreement 842,492 to S.R.; Newcastle University Academic Track (NUAcT) Fellowship to S.R.; Fulbright Research Scholarship to S.R.; Lister Institute Prize Fellowship to S.R.; Academy of Medical Sciences Springboard Award to S.R.; a grant from the Air Force Office of Scientific Research (AFOSR, FA9550‐23‐1‐0533) to S.R.; Leverhulme Research Leadership Award to S.R.; A.G. is supported by the AFOSR; A.R-G. is supported by a NUAcT PhD studentship.

\appendix
\section*{Appendix}
The Appendix contains additional information for the content presented in the main body.
\paragraph{Reproducibility and code availability.} All benchmarks are described in the Methods section of the manuscript. To enable the reproduction of all experiments, the relevant source code is publicly available at \url{https://github.com/alejandrorg00/ngm-sgd_project}.
\paragraph{Computer resources.} All experiments were run on a system with an Intel Core i9-14900K CPU (8 P-cores @ 3.2-6.0 GHz, 16 E-cores @ 2.4-4.4 GHz; 32 threads), 64 GB DDR5-4800 MHz RAM (2×32 GB, CL40), and a PNY NVIDIA RTX 4000 Ada Lovelace GPU (20 GB GDDR6). Code was implemented using Python 3.10.16 in PyTorch 2.5.1 with CUDA 11.8.
\paragraph{Code of ethics.} We used only publicly available datasets, and our work is confined to basic research without direct social impact. The authors declare no conflicts of interest.  
\section{Evaluation metrics}
\label{sec:metrics}
Throughout this work, we quantify model performance using \textbf{classification accuracy} -- the percentage of correctly predicted labels on a test set. Formally, for the test dataset $D$ and model $F_\theta$, we denote the classification accuracy $\mathbf{A}(D,F_\theta)$ as the percentage of samples in $D$ that are correctly classified by $F_\theta$. In practice, we evaluate on the full test set for the first task, although a reduced evaluation set of 1,000 examples per task, sampled uniformly at random in each run, has been shown to closely approximate the full-set results \citep{de_lange_continual_2022,hess_two_2023}.

\paragraph{Final average accuracy.} We further define the final average accuracy as the average classification accuracy measured at the end of training on each context. Formally, consider a continual‐learning sequence of $K$ contexts (tasks). Let $F_{\theta_{\mathrm{f}}}^{(k)}$ denote the model parameters immediately after training on context $k$, and let $D$ be its corresponding held‐out test set. Hence, we define
\begin{equation*}
    \textbf{avg-ACC} = \frac{1}{K} \sum_{k=1}^K \mathbf{A}(D,F_{\theta_{\mathrm{f}}}^{(k)}).
\end{equation*}
This metric captures overall end‐of‐sequence performance but does not reflect transient drops in accuracy that may occur during intermediate stages of training.

\paragraph{Average minimum accuracy.} To assess worst‐case retention within each context, we track the lowest classification accuracies observed for each context and define the average minimum accuracy as the average of these minimum accuracies. Formally, with $I_{\mathcal{C}}$ denoting the set of training‐iteration indices following completion of context $k$, we have
\begin{equation*}
    \textbf{avg-min-ACC} = \frac{1}{K-1} \sum_{k=2}^{K} \min_{i \in I_{\mathcal{C}}}  \mathbf{A}(D,F_{\theta_{i}}^{(k)}),
\end{equation*}
where $F_{\theta_{i}}^{(k)}$ indicates the model at the $i$-th training iteration in context $k$.

\paragraph{Worst-case accuracy.} To capture the balance between plasticity (learning the most recent task) and stability (preserving earlier tasks) \citep{duong_adaptive_2023, lapacz_exploring_2024}, we define worst-case accuracy as
\begin{equation*}
    \textbf{WC-ACC} = \frac{1}{K} \mathbf{A}(D,F_{\theta_{\mathrm{f}}}^{(K)}) + \left(1 - \frac{1}{K} \right) \textbf{avg-min-ACC}.
\end{equation*}
Hence, a high \textbf{WC-ACC} indicates both effective adaptation to new tasks and minimal forgetting of prior tasks. Note that we are evaluating this at the end of the experiment, but it can also be considered at each iteration.

\paragraph{Average stability gap.} We define the average stability gap as the mean relative drop in classification accuracy that occurs when shifting from one context to the next \citep{lapacz_exploring_2024}. For a sequence of $K$ contexts, the stability gap at the transition from context $k-1$ to context $k$ is given by (Table \ref{tab:stabgap_allmetrics})
\begin{equation*}
    \textbf{SG}_k = \frac{\mathbf{A}(D,F_{\theta_{\mathrm{f}}}^{(k-1)}) - \min_{i \in I_\mathcal{C}}  \mathbf{A}(D,F_{\theta_{i}}^{(k)})}{\mathbf{A}(D,F_{\theta_{\mathrm{f}}}^{(k-1)})},
\end{equation*}
and the average stability gap is then:
\begin{equation*}
    \textbf{avg-SG} = \frac{1}{K-1} \sum_{k=2}^K \textbf{SG}_k.
\end{equation*}
By normalizing each drop by the accuracy just before the transition, we ensure that stability gaps are directly comparable even when the baseline performances differ across tasks. A large \textbf{avg-SG} signals that the model experiences significant forgetting immediately after each context switch, underscoring its lack of robustness in continual learning settings.
\begin{table*}[t]
\centering
\caption{\textbf{Stability gap metrics.} For all joint training benchmarks, we report the per-transition stability gaps (mean $\pm$ std). Highlighted values indicate the best results.}
\label{tab:stabgap_allmetrics}
\centering
    \resizebox{\textwidth}{!}{
    \begin{tabular}{@{}cccccc@{}}
    \toprule
    & & \textbf{SG($T_{1 \to 2}$)} & \textbf{SG($T_{2 \to 3}$)} & \textbf{SG($T_{3 \to 4}$)} & \textbf{SG($T_{4 \to 5}$)} \\
    \midrule

    \multirow[c]{4}{*}{\shortstack{Split\\MNIST}}
    & NGM-SGD (ours) & \textbf{0.028 $\pm$ 0.015} & 0.012 $\pm$ 0.006 & 0.014 $\pm$ 0.008 & \textbf{0.013 $\pm$ 0.004} \\
    & MSGD     & 1.000 $\pm$ 0.000 & 0.032 $\pm$ 0.013 & 0.018 $\pm$ 0.006 & 0.019 $\pm$ 0.009 \\
    & Adam     & 0.883 $\pm$ 0.133 & 0.172 $\pm$ 0.062 & 0.052 $\pm$ 0.032 & 0.025 $\pm$ 0.013 \\
    & SGD      & 0.100 $\pm$ 0.115 & \textbf{0.009 $\pm$ 0.005} & \textbf{0.007 $\pm$ 0.006} & 0.019 $\pm$ 0.012 \\
    \midrule

    \multirow[c]{4}{*}{\shortstack{Split\\CIFAR-10}}
    & NGM-SGD (ours) & \textbf{0.133 $\pm$ 0.097} & \textbf{0.072 $\pm$ 0.060} & \textbf{0.072 $\pm$ 0.094} & \textbf{0.260 $\pm$ 0.090} \\
    & MSGD    & 0.476 $\pm$ 0.007 & 0.321 $\pm$ 0.089 & 0.116 $\pm$ 0.117 & 0.786 $\pm$ 0.095 \\
    & Adam    & 0.357 $\pm$ 0.103 & 0.148 $\pm$ 0.222 & 0.074 $\pm$ 0.054 & 0.661 $\pm$ 0.246 \\
    & SGD     & 0.389 $\pm$ 0.037 & 0.128 $\pm$ 0.112 & 0.120 $\pm$ 0.091 & 0.327 $\pm$ 0.184 \\
    \midrule

    \multirow[c]{4}{*}{\shortstack{Split\\mini-ImageNet}}
    & NGM-SGD (ours) & \textbf{0.515 $\pm$ 0.078} & 0.343 $\pm$ 0.088 & \textbf{0.213 $\pm$ 0.162} & \textbf{0.128 $\pm$ 0.202} \\
    & MSGD    & 0.786 $\pm$ 0.029 & 0.310 $\pm$ 0.204 & 0.320 $\pm$ 0.101 & 0.221 $\pm$ 0.141 \\
    & Adam    & 0.587 $\pm$ 0.133 & 0.507 $\pm$ 0.214 & 0.284 $\pm$ 0.093 & 0.157 $\pm$ 0.156 \\
    & SGD     & 0.636 $\pm$ 0.110 & \textbf{0.305 $\pm$ 0.259} & 0.245 $\pm$ 0.169 & 0.141 $\pm$ 0.212 \\
    \midrule

    \multirow[c]{4}{*}{\shortstack{Rotated\\MNIST}}
    & NGM-SGD (ours) & 0.050 $\pm$ 0.017 & \textbf{0.057 $\pm$ 0.025} & - & - \\
    & MSGD    & 0.097 $\pm$ 0.021 & 0.138 $\pm$ 0.020 & - & - \\
    & Adam    & 0.085 $\pm$ 0.020 & 0.099 $\pm$ 0.010 & - & - \\
    & SGD     & \textbf{0.045 $\pm$ 0.033} & 0.062 $\pm$ 0.026 & - & - \\
    \midrule

    \multirow[c]{4}{*}{\shortstack{Domain\\CIFAR-100}}
    & NGM-SGD (ours) & 0.334 $\pm$ 0.070 & \textbf{0.114 $\pm$ 0.047} & - & - \\
    & MSGD    & \textbf{0.329 $\pm$ 0.063} & 0.135 $\pm$ 0.038 & - & - \\
    & Adam    & 0.370 $\pm$ 0.062 & 0.142 $\pm$ 0.049 & - & - \\
    & SGD     & 0.349 $\pm$ 0.087 & 0.137 $\pm$ 0.057 & - & - \\
    \bottomrule
    \end{tabular}
    }
\end{table*}
\paragraph{Cumulative and new-task accuracy.} In addition to evaluating retention on the first task, we evaluate performance on dynamically defined test sets as new contexts are introduced. At context $k$, we define the cumulative test set as $D_{\mathrm{cum}}^{(k)} = \bigcup_{j=1}^{k} D_j$, and compute the cumulative accuracy as $\mathbf{A}(D_{\mathrm{cum}}^{(k)}, F_{\theta_i}^{(k)})$. Conversely, new-task accuracy is evaluated exclusively on the test set of the currently learned context, $\mathbf{A}(D_k, F_{\theta_i}^{(k)})$. We refer to cumulative accuracy measured at the end of training on context $k$, $\mathbf{A}(D_{\mathrm{cum}}^{(k)}, F_{\theta_{\mathrm{f}}}^{(k)})$, as end-cumulative accuracy.
\section{Hyperparameter search} \label{sec:hyperparams}
\begin{figure}[pos=!htbp]
  \centering
  \includegraphics[width=\columnwidth]{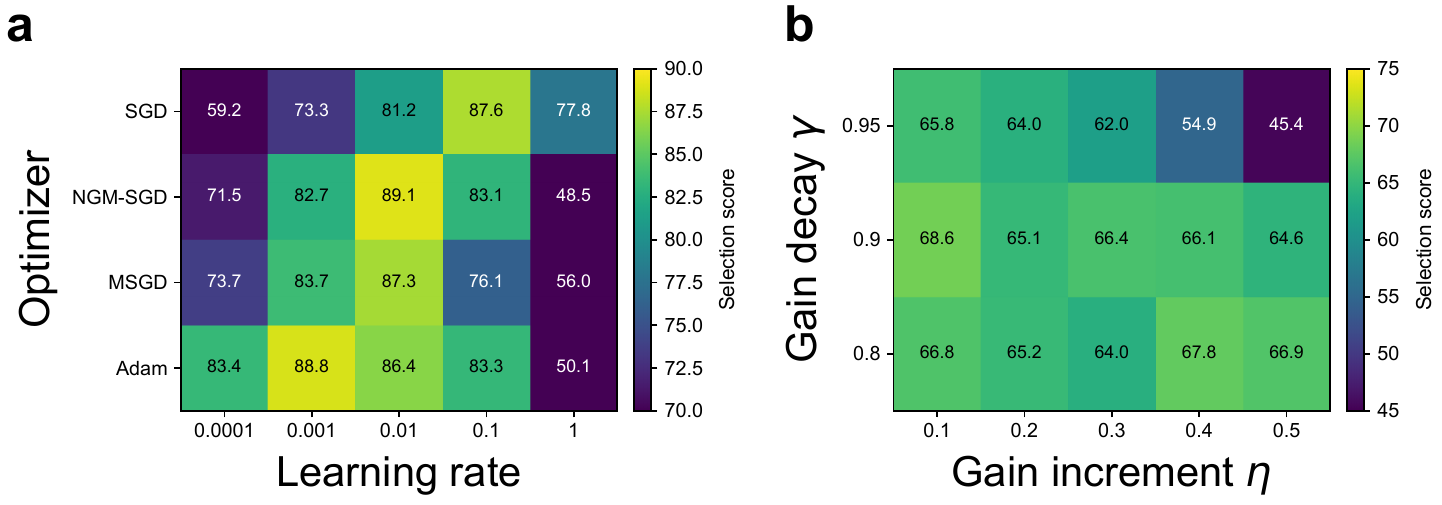}
\caption{\textbf{Hyperparameter sweep on Split CIFAR-10.} \textbf{a)} Scores on best accuracy with the learning rate sweep for all methods. \textbf{b)} Scores on best accuracy and least stability gap with the hyperparameter sweep of $(\gamma,\eta)$ over the NGM-SGD method.}
  \label{fig:app_sweep_cifar}
\end{figure}

\begin{table*}[!ht]
\caption{\textbf{Hyperparameter search:} Final accuracies (ACC) for all tasks in the Split CIFAR-10 benchmark comparing different learning rates.}
  \centering
  \resizebox{\textwidth}{!}{
  \begin{tabular}{@{}ccccccc@{}}
  \toprule
    \textbf{LR} &  & \textbf{T1} & \textbf{T2} & \textbf{T3} & \textbf{T4} & \textbf{T5} \\
    \midrule
    \multirow[c]{4}{*}{\shortstack{1e-4}}
      & NGM-SGD (ours) & 83.12 $\pm$ 1.35 & 74.14 $\pm$ 1.75 & 75.68 $\pm$ 2.38 & 75.83 $\pm$ 2.17 & 54.95 $\pm$ 5.02 \\
    & MSGD & 80.40 $\pm$ 1.09 & 74.89 $\pm$ 0.74 & 77.15 $\pm$ 1.00 & 77.33 $\pm$ 0.67 & 61.42 $\pm$ 1.78 \\
    & Adam & 92.07 $\pm$ 0.67 & 84.16 $\pm$ 1.89 & 87.53 $\pm$ 0.87 & 84.55 $\pm$ 2.65 & 75.02 $\pm$ 5.24 \\
    & SGD & 55.99 $\pm$ 4.80 & 64.75 $\pm$ 2.75 & 65.12 $\pm$ 1.73 & 60.76 $\pm$ 1.95 & 55.75 $\pm$ 2.35 \\
    \midrule
    \multirow[c]{4}{*}{\shortstack{1e-3}}
      & NGM-SGD (ours) & 91.09 $\pm$ 1.46 & 85.89 $\pm$ 2.82 & 85.84 $\pm$ 1.50 & 85.77 $\pm$ 1.44 & 70.84 $\pm$ 4.72 \\
    & MSGD & 91.36 $\pm$ 1.35 & 85.31 $\pm$ 1.77 & 85.76 $\pm$ 1.22 & 86.24 $\pm$ 1.12 & 73.72 $\pm$ 2.82 \\
    & Adam & 96.25 $\pm$ 0.64 & 91.10 $\pm$ 5.16 & 90.91 $\pm$ 2.40 & 88.79 $\pm$ 3.09 & 84.66 $\pm$ 4.48 \\
    & SGD & 80.43 $\pm$ 0.91 & 73.61 $\pm$ 1.62 & 76.84 $\pm$ 1.06 & 77.55 $\pm$ 0.87 & 60.59 $\pm$ 1.45 \\
    \midrule
    \multirow[c]{4}{*}{\shortstack{1e-2}}
      & NGM-SGD (ours) & 95.95 $\pm$ 0.60 & 91.28 $\pm$ 2.09 & 91.65 $\pm$ 1.11 & 87.81 $\pm$ 3.99 & 86.46 $\pm$ 6.34 \\
    & MSGD & 96.31 $\pm$ 0.45 & 91.05 $\pm$ 2.63 & 90.55 $\pm$ 2.12 & 88.83 $\pm$ 5.95 & 80.43 $\pm$ 8.17 \\
    & Adam & 93.32 $\pm$ 1.38 & 84.87 $\pm$ 4.67 & 88.99 $\pm$ 3.07 & 88.65 $\pm$ 5.37 & 84.46 $\pm$ 2.38 \\
    & SGD & 89.95 $\pm$ 1.62 & 86.44 $\pm$ 2.91 & 85.64 $\pm$ 1.70 & 86.46 $\pm$ 1.40 & 64.28 $\pm$ 5.37 \\
    \midrule
    \multirow[c]{4}{*}{\shortstack{1e-1}}
      & NGM-SGD (ours) & 90.71 $\pm$ 1.69 & 83.58 $\pm$ 3.40 & 83.88 $\pm$ 2.60 & 84.52 $\pm$ 5.18 & 81.02 $\pm$ 4.46 \\
    & MSGD & 93.68 $\pm$ 2.50 & 79.86 $\pm$ 4.30 & 80.75 $\pm$ 5.71 & 82.93 $\pm$ 4.75 & 61.79 $\pm$ 16.02 \\
    & Adam & 90.77 $\pm$ 2.10 & 85.28 $\pm$ 5.04 & 85.70 $\pm$ 5.31 & 89.58 $\pm$ 2.30 & 76.65 $\pm$ 8.16 \\
    & SGD & 94.45 $\pm$ 1.60 & 89.42 $\pm$ 3.78 & 90.97 $\pm$ 4.94 & 88.06 $\pm$ 6.46 & 84.49 $\pm$ 2.84 \\
    \midrule
    \multirow[c]{4}{*}{\shortstack{1}}
      & NGM-SGD (ours) & 61.03 $\pm$ 12.49 & 54.06 $\pm$ 5.41 & 52.60 $\pm$ 9.35 & 51.25 $\pm$ 12.23 & 46.09 $\pm$ 8.87 \\
    & MSGD & 79.44 $\pm$ 3.71 & 58.78 $\pm$ 12.59 & 56.10 $\pm$ 12.79 & 56.02 $\pm$ 11.34 & 53.52 $\pm$ 10.25 \\
    & Adam & 82.69 $\pm$ 3.64 & 53.35 $\pm$ 11.77 & 46.58 $\pm$ 9.12 & 50.49 $\pm$ 0.98 & 40.96 $\pm$ 18.08 \\
    & SGD & 88.02 $\pm$ 1.65 & 80.58 $\pm$ 4.27 & 81.15 $\pm$ 3.19 & 83.17 $\pm$ 2.11 & 71.51 $\pm$ 14.14 \\
    \bottomrule
    \label{tab:lr_metrics}
  \end{tabular}
  }
\end{table*}
We conducted a systematic hyperparameter search using CIFAR-10 as a representative middle-ground benchmark. CIFAR-10 strikes a balance between easier datasets such as MNIST and more challenging ones like CIFAR-100 and mini-ImageNet, making it an appropriate testbed for controlled comparisons. For all optimizers under consideration, we performed a learning rate sweep over \{1.0, 0.1, 0.01, 0.001, 0.0001\}. In addition, for NGM-SGD we searched over modulation parameters, exploring $\gamma \in \{0.8, 0.9, 0.95\}$ and $\eta \in \{0.1, 0.2, 0.3, 0.4, 0.5\}$. We note that excessively high gains are both biologically implausible \citep{munn_neuronal_2023}, as they lead to saturation, and practically undesirable, as they introduce instability into training.
\paragraph{NGM-SGD hyperparameters.} The NGM-SGD method presents three parameters: the gain baseline $g_0$, which we set to $1$ since its effect can be absorbed into the learning rate, the decay $\gamma$, and the phasic bump $\eta$. To explore which parameter values fit best for CIFAR-10, we computed a score that jointly accounts for accuracy and stability (Table~\ref{tab:ge_metrics}). In particular, we considered an effective accuracy that penalizes variability, and a stability gap term that penalizes both its magnitude and variability ($\text{score}_{\text{NGM}} = \big(\text{avg-ACC} - \text{std-ACC}\big) - 100 \cdot \big(\text{avg-SG} + \text{std-SG}\big)$). The best-performing combination was $(\gamma, \eta) = (0.9, 0.2)$ (Figure~\ref{fig:app_sweep_cifar}b). While the choice of $\gamma = 0.9$ has proven consistently robust across all other tasks, the value of $\eta$ requires tuning depending on task difficulty. Higher values of $\eta$ can be used for easier datasets, whereas lower values are preferable for more complex datasets to avoid instabilities when uncertainty is higher.
\paragraph{Learning rate search.} To identify the most effective learning rates, we evaluated end-task performance (Table~\ref{tab:lr_metrics}) using a scoring criterion that rewards accuracy. Specifically, we considered the effective accuracy, defined as the average accuracy penalized by its variability, which favors configurations that achieve both high performance and robustness ($\text{score}_{\text{LR}} = \text{avg-ACC} - \text{std-ACC})$. This analysis revealed that the best values are $0.1$ for SGD, $0.01$ for MSGD and NGM-SGD, and $0.001$ for Adam (Figure~\ref{fig:app_sweep_cifar}a). These settings consistently yielded the best performance across datasets, except Rotated MNIST, where MSGD achieved higher accuracy with a learning rate of $0.1$.

\begin{table}[t]
  \centering
  \caption {\textbf{Hyperparameter search:} Average final accuracies (avg-ACC) and average stability gaps (avg-SG) in the Split CIFAR-10 benchmark comparing different combinations of the $(\gamma, \eta)$ parameters in the NGM-SGD method.}
\begin{tabular}{@{}cccc@{}}
\toprule
 $\gamma$ &  $\eta$ & \textbf{avg-ACC} ($\uparrow$) & \textbf{avg-SG} ($\downarrow$) \\
\midrule
     0.80 &     0.1 &            90.134 $\pm$ 1.292 &              0.182 $\pm$ 0.038 \\
     0.80 &     0.2 &            90.920 $\pm$ 1.396 &              0.182 $\pm$ 0.062 \\
     0.80 &     0.3 &            91.678 $\pm$ 0.875 &              0.215 $\pm$ 0.053 \\
     0.80 &     0.4 &            91.220 $\pm$ 1.068 &              0.176 $\pm$ 0.047 \\
     0.80 &     0.5 &            90.198 $\pm$ 1.471 &              0.170 $\pm$ 0.049 \\
     0.90 &     0.1 &            91.098 $\pm$ 1.263 &              0.159 $\pm$ 0.054 \\
     0.90 &     0.2 &            90.630 $\pm$ 1.576 &              0.134 $\pm$ 0.044 \\
     0.90 &     0.3 &            90.550 $\pm$ 1.751 &              0.171 $\pm$ 0.053 \\
     0.90 &     0.4 &            90.476 $\pm$ 0.848 &              0.178 $\pm$ 0.057 \\
     0.90 &     0.5 &            90.156 $\pm$ 1.256 &              0.201 $\pm$ 0.043 \\
     0.95 &     0.1 &            90.094 $\pm$ 0.966 &              0.191 $\pm$ 0.043 \\
     0.95 &     0.2 &            89.516 $\pm$ 1.728 &              0.190 $\pm$ 0.047 \\
     0.95 &     0.3 &            91.156 $\pm$ 1.190 &              0.242 $\pm$ 0.038 \\
     0.95 &     0.4 &            90.846 $\pm$ 1.186 &              0.237 $\pm$ 0.110 \\
     0.95 &     0.5 &            89.042 $\pm$ 1.676 &              0.325 $\pm$ 0.094 \\
\bottomrule
\label{tab:ge_metrics}
\end{tabular}
\end{table}

\section{Gain dynamics}
\label{app:gain_dynamics}
In the main text, we model task-dependent modulation of plasticity through a neuron-specific scalar gain $g_i(t)$ that multiplicatively scales the incoming weights of each output neuron $i$. The gain evolves according to the discrete-time update provided by Equation~\eqref{eq:gain_update} where $g_0$ is a basal gain level, $H(y_t)$ is the entropy of the model's output at time $t$, $\gamma \in (0,1)$ controls the decay of the gain toward baseline, and $\eta > 0$ scales the impact of the uncertainty signal. This dynamic is motivated by experimental findings showing that noradrenergic arousal reports surprise and modulates neuronal gain during perceptual updating \citep{wainstein_gain_2025, jordan_locus_2024}. In particular, \citet{wainstein_gain_2025} model phasic gain increases and their subsequent tonic decay using a first-order linear differential equation driven by uncertainty, a structure closely mirrored by our update rule. In what follows, we derive the corresponding continuous-time form to make this connection explicit, and then clarify how the discrete-time equation relates to a classical exponential moving average (EMA) -- a formulation more familiar in computing contexts.

\subsection{Continuous-time gain}
We will derive the continuous-time form of the gain update starting from Equation~\eqref{eq:gain_update}. First, we subtract $g(t)$ from both sides, i.e. 
\begin{align*}
    g(t+1) - g(t) &= \gamma \; g(t) + (1-\gamma) g_0 + \eta \; H(y) - g(t) \\
    &= \left( \gamma - 1 \right) \; g(t) + (1-\gamma) g_0 + \eta \; H(y) \\
    &= \left( 1 -\gamma \right) \left(g_0 - g(t) \right) + \eta \; H(y).
\end{align*} 
Next, dividing both sides by the discrete time interval $\Delta t$, we obtain
\begin{equation*}
    \frac{g(t+1) - g(t)}{\Delta t} = \frac{\left( 1 -\gamma \right)}{\Delta t} \left(g_0 - g(t) \right) + \frac{\eta}{\Delta t} H(y).
\end{equation*} 
Considering Euler's method for numerical integration and defining the characteristic time constant $\tau = \Delta t / \left( 1 -\gamma \right)$, and the coefficient $\kappa = \left( \eta/\Delta t \right) \tau = \eta / \left( 1 -\gamma \right) $, we arrive at the continuous-time differential equation
\begin{equation}
    \tau \frac{dg(t)}{dt} = \left(g_0 - g(t) \right) + \kappa \; H(y)
\end{equation} 
This continuous-time formulation makes explicit that $g(t)$ behaves as a low-pass filter of the entropy signal $H(y_t)$ around the basal gain level $g_0$, with time constant $\tau$. Transient increases in uncertainty drive $g(t)$ above baseline, followed by a gradual relaxation back to $g_0$, reminiscent of noradrenergic bursts and their subsequent decay in arousal-related signals, such as pupil dilation, during perceptual transitions and unexpected events \citep{wainstein_gain_2025, sales2019}.

\subsection{Relation to exponential moving averages}
\begin{figure*}[pos=!htbp]
  \centering
  \includegraphics[width=\textwidth]{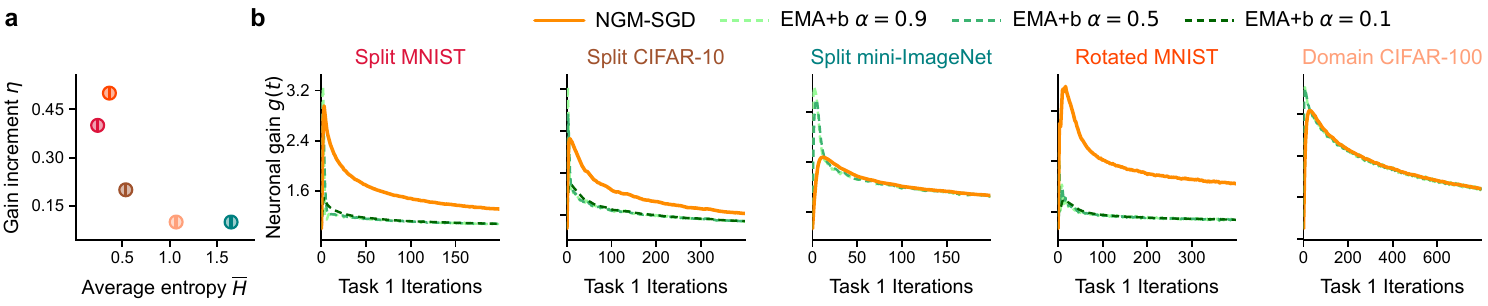}
  \caption{\textbf{Gain dynamics:} \textbf{a)} Relationship between the average output entropy of each dataset and the corresponding gain scale. \textbf{b)} Gain trajectories (Task 1) comparing NGM-SGD with EMA-based baselines for the domain- and class-incremental benchmarks. EMA (+ baseline) settings match only the special case $\eta = 1-\gamma$, whereas NGM-SGD allows independent control of decay and entropy-driven modulation.}
  \label{fig:app_gainEMA_H}
\end{figure*}
The discrete-time gain update in Equation~\eqref{eq:gain_update} is reminiscent of an EMA conditioned on entropy. Here, we make this connection explicit and clarify the differences with a standard EMA. A canonical EMA of a signal $x_t$ around an equilibrium level $g_0$ is typically written as
\begin{equation}
    g_{\mathrm{EMA}}(t+1)
    = (1-\alpha)\, g_{\mathrm{EMA}}(t) + \alpha \bigl[g_0 + x_t\bigr],
    \qquad \alpha \in (0,1),
    \label{eq:ema_standard}
\end{equation}
where $\alpha \in (0,1)$ sets the smoothing factor in the exponential moving average. If we choose $x_t = H(y_t)$ and expand Equation~\eqref{eq:ema_standard}, we obtain
\begin{equation*}
    g_{\mathrm{EMA}}(t+1)
    = (1-\alpha)\, g_{\mathrm{EMA}}(t) + \alpha g_0 + \alpha H(y_t).
\end{equation*}
Comparing this to the NGM-SGD update in Equation~\eqref{eq:gain_update} shows that the two dynamics coincide if and only if
\begin{equation*}
    \gamma = 1 - \alpha
    \quad\text{and}\quad
    \eta = \alpha.
    \label{eq:ema_identification}
\end{equation*}
In other words, when $\eta = 1-\gamma$, we can rewrite our gain update exactly as an EMA of the entropy signal around the baseline $g_0$. This special case makes precise in what sense our formulation ``recalls'' an EMA with a baseline term -- as happens with the cases of Split mini-ImageNet and Domain CIFAR-100 (see Figure~\ref{fig:app_gainEMA_H}b). Hence,  the single parameter $\alpha$ plays a dual role: it determines both the temporal decay factor $(1-\alpha)$ and the scaling of the new signal $\alpha H(y_t)$. Increasing $\alpha$ makes the process more reactive but also less smooth; decreasing $\alpha$ increases smoothness but suppresses the impact of new information as shown in Figure~\ref{fig:app_gainEMA_H}b. In contrast, in Equation~\eqref{eq:gain_update} the decay of the gain is governed solely by $\gamma$, while the influence of the entropy signal is controlled by $\eta$. This decoupling allows us, for example, to choose a large $\gamma$ (slow decay, strong stability) together with a large $\eta$ (large gain bursts at task transitions), leading to a better attenuation of the stability gap\footnote{For instance, Table~\ref{tab:ge_metrics} shows how combinations of $( \gamma, \eta )$ of EMA-equivalent settings $\{(0.9, 0.1); (0.8, 0.2)\}$ are outperformed by the case $(0.9, 0.2)$, which lies strictly outside the manifold $\eta = 1-\gamma$.}.

Furthermore, this decoupling is essential in practice: across benchmarks we keep $\gamma = 0.9$ to ensure slow decay and strong stability, while tuning $\eta$ to reflect the entropy structure of each dataset. Datasets with lower overall entropy require larger $\eta$ to induce sufficiently large gain transients and achieve measurable reductions in the stability gap, whereas more complex datasets with inherently higher entropy call for smaller $\eta$ to keep the gain dynamics within a bounded regime, as illustrated in Figure~\ref{fig:app_gainEMA_H}a.
\section{Momentum resetting in multi-timescale optimizers}
\label{sec:app_resetting}
To further examine whether momentum-driven inertia contributes to the stability gap, we repeated our experiments across all benchmarks using a modified version of Adam and MSGD in which the internal momentum buffers were cleared at every task change. This was implemented by providing a task-change signal that forces the optimizer to clear its internal state, thereby removing all previously accumulated velocity -- an approach that is not task-agnostic, unlike neuromodulatory gain (our NGM-SGD method). As shown in Table~\ref{tab:reset_metrics}, momentum resets indeed reduce part of the instability in most cases (see highlighted colors), confirming that accumulated velocity can overshoot under abrupt loss-landscape shifts. However, even with perfect knowledge of task switches, these reset variants remain less stable than NGM-SGD across benchmarks.

\begin{table*}[h!]
  \centering
  \caption{\textbf{Oracle reset results across benchmarks.} Main quantitative metrics for all optimizers across benchmarks under joint training with momentum resets. Bold values indicate the best result per column. For Adam and MSGD, color highlights denote the metrics in which the momentum-reset variant outperforms its non-reset counterpart (\textcolor{OliveGreen}{Green} for Adam Reset and \textcolor{Blue}{Blue} for MSGD Reset).}
  \label{tab:reset_metrics}
  \centering
  \resizebox{\textwidth}{!}{
  \begin{tabular}{@{}cccccc@{}}
    \toprule
     &  & \textbf{avg-ACC ($\uparrow$)} & \textbf{avg-min-ACC ($\uparrow$)} & \textbf{WC-ACC ($\uparrow$)} & \textbf{avg-SG ($\downarrow$)} \\
    \midrule 
    \multirow[c]{6}{*}{\shortstack{Split\\MNIST}}
      & NGM-SGD (ours) & 99.015 $\pm$ 0.047 & \textbf{97.570 $\pm$ 0.465} & \textbf{97.696 $\pm$ 0.373} & \textbf{0.017 $\pm$ 0.005} \\
      & SGD & 98.975 $\pm$ 0.099 & 95.887 $\pm$ 2.887 & 96.293 $\pm$ 2.311 & 0.034 $\pm$ 0.029 \\
      & Adam & \textbf{99.088 $\pm$ 0.108} & 71.047 $\pm$ 3.753 & 76.537 $\pm$ 3.003 & 0.283 $\pm$ 0.038 \\
      & Adam Reset & 99.075 $\pm$ 0.147 & \textcolor{OliveGreen}{86.563 $\pm$ 6.869} & \textcolor{OliveGreen}{88.921 $\pm$ 5.496} & \textcolor{OliveGreen}{0.127 $\pm$ 0.069} \\
      & MSGD & 98.971 $\pm$ 0.086 & 72.518 $\pm$ 0.403 & 77.628 $\pm$ 0.329 & 0.267 $\pm$ 0.004 \\
      & MSGD Reset & 98.939 $\pm$ 0.100 & \textcolor{Blue}{72.823 $\pm$ 0.374} & \textcolor{Blue}{77.876 $\pm$ 0.305} & \textcolor{Blue}{0.264 $\pm$ 0.004} \\
    \midrule
    \multirow[c]{6}{*}{\shortstack{Split\\CIFAR-10}}
      & NGM-SGD (ours) & 90.630 $\pm$ 1.576 & \textbf{79.485 $\pm$ 3.875} & \textbf{80.880 $\pm$ 3.349} & \textbf{0.134 $\pm$ 0.043} \\
      & SGD & 89.478 $\pm$ 1.909 & 68.712 $\pm$ 4.958 & 71.868 $\pm$ 4.007 & 0.241 $\pm$ 0.059 \\
      & Adam & 90.342 $\pm$ 1.579 & 63.450 $\pm$ 7.839 & 67.692 $\pm$ 6.335 & 0.310 $\pm$ 0.088 \\
      & Adam Reset & \textbf{\textcolor{OliveGreen}{90.734 $\pm$ 1.447}} & \textcolor{OliveGreen}{74.890 $\pm$ 4.980} & \textcolor{OliveGreen}{76.302 $\pm$ 4.134} & \textcolor{OliveGreen}{0.193 $\pm$ 0.054} \\
      & MSGD & 89.434 $\pm$ 2.134 & 52.828 $\pm$ 3.898 & 58.348 $\pm$ 3.521 & 0.425 $\pm$ 0.044 \\
      & MSGD Reset & \textcolor{Blue}{89.976 $\pm$ 1.175} & \textcolor{Blue}{70.492 $\pm$ 5.190} & \textcolor{Blue}{73.118 $\pm$ 4.177} & \textcolor{Blue}{0.230 $\pm$ 0.058} \\
    \midrule
    \multirow[c]{6}{*}{\shortstack{Split\\mini-ImageNet}}
      & NGM-SGD (ours) & \textbf{48.312 $\pm$ 2.289} & \textbf{33.165 $\pm$ 2.435} & \textbf{36.040 $\pm$ 2.058} & \textbf{0.300 $\pm$ 0.071} \\
      & SGD & 42.988 $\pm$ 2.549 & 28.345 $\pm$ 3.671 & 30.288 $\pm$ 3.309 & 0.332 $\pm$ 0.098 \\
      & Adam & 47.208 $\pm$ 2.044 & 28.300 $\pm$ 3.371 & 32.052 $\pm$ 2.768 & 0.384 $\pm$ 0.077 \\
      & Adam Reset & 46.760 $\pm$ 1.903 & 20.690 $\pm$ 4.813 & 25.264 $\pm$ 3.949 & 0.561 $\pm$ 0.106 \\
      & MSGD & 46.824 $\pm$ 2.176 & 26.775 $\pm$ 2.417 & 30.416 $\pm$ 2.029 & 0.409 $\pm$ 0.067 \\
      & MSGD Reset & 44.372 $\pm$ 2.678 & \textcolor{Blue}{28.925 $\pm$ 2.337} & \textcolor{Blue}{31.176 $\pm$ 2.162} & \textcolor{Blue}{0.335 $\pm$ 0.076} \\
    \midrule
    \multirow[c]{6}{*}{\shortstack{Rotated\\MNIST}}
      & NGM-SGD (ours) & 94.948 $\pm$ 0.121 & \textbf{90.015 $\pm$ 1.451} & \textbf{91.542 $\pm$ 0.973} & 0.054 $\pm$ 0.015 \\
      & SGD & 94.139 $\pm$ 0.255 & 89.107 $\pm$ 1.935 & 90.787 $\pm$ 1.300 & \textbf{0.053 $\pm$ 0.021} \\
      & Adam & \textbf{95.777 $\pm$ 0.195} & 87.177 $\pm$ 1.060 & 89.910 $\pm$ 0.721 & 0.092 $\pm$ 0.011 \\
      & Adam Reset & 95.545 $\pm$ 0.235 & \textcolor{OliveGreen}{88.499 $\pm$ 1.073} & \textcolor{OliveGreen}{90.795 $\pm$ 0.727} & \textcolor{OliveGreen}{0.075 $\pm$ 0.012} \\
      & MSGD & 95.297 $\pm$ 0.217 & 84.329 $\pm$ 1.359 & 87.823 $\pm$ 0.914 & 0.117 $\pm$ 0.014 \\
      & MSGD Reset & \textcolor{Blue}{95.329 $\pm$ 0.216} & \textcolor{Blue}{85.902 $\pm$ 1.918} & \textcolor{Blue}{88.880 $\pm$ 1.282} & \textcolor{Blue}{0.101 $\pm$ 0.020} \\
    \midrule  
    \multirow[c]{6}{*}{\shortstack{Domain\\CIFAR-100}}
      & NGM-SGD (ours) & 68.700 $\pm$ 0.881 & \textbf{53.420 $\pm$ 2.804} & \textbf{58.290 $\pm$ 1.916} & 0.224 $\pm$ 0.042 \\
      & SGD & 65.463 $\pm$ 1.204 & 49.050 $\pm$ 3.177 & 54.950 $\pm$ 2.229 & 0.243 $\pm$ 0.052 \\
      & Adam & 69.163 $\pm$ 0.871 & 51.615 $\pm$ 2.653 & 57.323 $\pm$ 1.840 & 0.256 $\pm$ 0.040 \\
      & Adam Reset & \textbf{\textcolor{OliveGreen}{69.823 $\pm$ 1.111}} & 49.415 $\pm$ 3.210 & 55.970 $\pm$ 2.201 & 0.295 $\pm$ 0.048 \\
      & MSGD & 67.160 $\pm$ 0.994 & 51.370 $\pm$ 2.234 & 56.867 $\pm$ 1.556 & 0.232 $\pm$ 0.037 \\
      & MSGD Reset & \textcolor{Blue}{67.463 $\pm$ 1.075} & \textcolor{Blue}{52.350 $\pm$ 2.248} & \textcolor{Blue}{57.943 $\pm$ 1.567} & \textbf{\textcolor{Blue}{0.214 $\pm$ 0.038}} \\
    \bottomrule
  \end{tabular}
  }
\end{table*}
\section{Benchmarks under partial replay}
\label{app:replay}
To evaluate our optimizer in a more realistic scenario, we also consider experiments in an experience replay (ER) setting \citep{rolnick2019experiencereplaycontinuallearning}. In this regime, each dataset is equipped with a finite replay buffer that is class-balanced and capped at a fixed size. After each task, we generate the buffer in a class-balanced manner, and during training, we construct each batch as a homogeneous mixture of current-task examples and replayed examples. This ER protocol ensures that the replay stream is not biased toward the most recent task, all classes remain represented, and optimizers are compared under the same rehearsal conditions.
\begin{table*}[t]
  \centering
  \caption{\textbf{Experience replay results across all benchmarks.} We report the main quantitative metrics under different replay-buffer sizes. Each method is evaluated with class-balanced replay and mixed batches of new and replayed samples. Bold values indicate the best result per column. For Adam and MSGD, color highlights denote the metrics in which the momentum-reset variant outperforms its non-reset counterpart (\textcolor{OliveGreen}{Green} for Adam Reset and \textcolor{Blue}{Blue} for MSGD Reset).}
  \label{tab:ER_metrics}
  \begin{adjustbox}{
      max width=\textwidth,
      max totalheight=0.9\textheight,
      keepaspectratio
  }
  \centering
  \resizebox{\textwidth}{!}{
  \begin{tabular}{@{}cccccc@{}}
    \toprule
     &  & \textbf{avg-ACC ($\uparrow$)} & \textbf{avg-min-ACC ($\uparrow$)} & \textbf{WC-ACC ($\uparrow$)} & \textbf{avg-SG ($\downarrow$)} \\
    \midrule 
    \multirow[c]{6}{*}{\shortstack{Split\\MNIST\\{\footnotesize \textbf{(1K buffer)}}}}
      & NGM-SGD (ours) & \textbf{98.321 $\pm$ 0.182} & \textbf{95.123 $\pm$ 1.219} & \textbf{95.344 $\pm$ 0.988} & \textbf{0.038 $\pm$ 0.012} \\
      & SGD & 97.991 $\pm$ 0.319 & 94.251 $\pm$ 1.813 & 94.517 $\pm$ 1.470 & 0.044 $\pm$ 0.018 \\
      & Adam & 97.974 $\pm$ 0.259 & 48.515 $\pm$ 5.823 & 57.973 $\pm$ 4.662 & 0.505 $\pm$ 0.059 \\
      & Adam Reset & 97.918 $\pm$ 0.175 & \textcolor{OliveGreen}{81.849 $\pm$ 8.599} & \textcolor{OliveGreen}{84.700 $\pm$ 6.880} & \textcolor{OliveGreen}{0.166 $\pm$ 0.086} \\
      & MSGD & 97.927 $\pm$ 0.249 & 52.712 $\pm$ 3.238 & 61.280 $\pm$ 2.600 & 0.462 $\pm$ 0.033 \\
      & MSGD Reset & \textcolor{Blue}{97.952 $\pm$ 0.103} & \textcolor{Blue}{55.560 $\pm$ 3.841} & \textcolor{Blue}{63.597 $\pm$ 3.073} & \textcolor{Blue}{0.433 $\pm$ 0.039} \\
    \midrule
    \multirow[c]{6}{*}{\shortstack{Split\\CIFAR-10\\{\footnotesize \textbf{(1K buffer)}}}}
      & NGM-SGD (ours) & 62.788 $\pm$ 1.521 & \textbf{50.372 $\pm$ 2.887} & 43.170 $\pm$ 2.374 & \textbf{0.334 $\pm$ 0.044} \\
      & SGD & 63.936 $\pm$ 1.798 & 40.445 $\pm$ 4.002 & 35.506 $\pm$ 3.374 & 0.483 $\pm$ 0.057 \\
      & Adam & \textbf{69.820 $\pm$ 2.930} & 46.098 $\pm$ 4.573 & 42.640 $\pm$ 4.346 & 0.431 $\pm$ 0.060 \\
      & Adam Reset & 64.944 $\pm$ 3.043 & \textcolor{OliveGreen}{48.410 $\pm$ 4.164} & \textcolor{OliveGreen}{\textbf{44.334 $\pm$ 3.506}} & \textcolor{OliveGreen}{0.350 $\pm$ 0.076} \\
      & MSGD & 65.576 $\pm$ 1.986 & 35.435 $\pm$ 5.486 & 31.834 $\pm$ 4.447 & 0.568 $\pm$ 0.075 \\
      & MSGD Reset & \textcolor{Blue}{66.312 $\pm$ 2.893} & \textcolor{Blue}{39.103 $\pm$ 3.021} & \textcolor{Blue}{35.240 $\pm$ 2.780} & \textcolor{Blue}{0.522 $\pm$ 0.045} \\
    \midrule
    \multirow[c]{6}{*}{\shortstack{Split\\CIFAR-10\\{\footnotesize \textbf{(3K buffer)}}}}
      & NGM-SGD (ours) & 76.722 $\pm$ 1.795 & \textbf{63.862 $\pm$ 5.975} & \textbf{60.326 $\pm$ 4.833} & \textbf{0.248 $\pm$ 0.077} \\
      & SGD & 73.652 $\pm$ 3.375 & 49.898 $\pm$ 7.246 & 47.862 $\pm$ 6.113 & 0.394 $\pm$ 0.096 \\
      & Adam & 79.076 $\pm$ 2.514 & 50.398 $\pm$ 6.527 & 51.322 $\pm$ 5.559 & 0.412 $\pm$ 0.081 \\
      & Adam Reset & \textcolor{OliveGreen}{\textbf{80.856 $\pm$ 2.313}} & \textcolor{OliveGreen}{50.898 $\pm$ 4.780} & \textcolor{OliveGreen}{51.728 $\pm$ 3.968} & 0.422 $\pm$ 0.058 \\
      & MSGD & 77.644 $\pm$ 2.921 & 34.215 $\pm$ 8.591 & 37.202 $\pm$ 7.282 & 0.613 $\pm$ 0.101 \\
      & MSGD Reset & 76.974 $\pm$ 2.489 & \textcolor{Blue}{36.532 $\pm$ 6.987} & \textcolor{Blue}{39.122 $\pm$ 5.675} & \textcolor{Blue}{0.577 $\pm$ 0.086} \\
    \midrule
    \multirow[c]{6}{*}{\shortstack{Split\\mini-ImageNet\\{\footnotesize \textbf{(1K buffer)}}}}
      & NGM-SGD (ours) & \textbf{23.760 $\pm$ 1.236} & \textbf{12.300 $\pm$ 1.442} & \textbf{11.308 $\pm$ 1.171} & 0.503 $\pm$ 0.055 \\
      & SGD & 19.044 $\pm$ 2.201 & 10.345 $\pm$ 2.036 & 9.492 $\pm$ 1.740 & \textbf{0.485 $\pm$ 0.110} \\
      & Adam & 23.236 $\pm$ 2.293 & 11.500 $\pm$ 1.181 & 10.824 $\pm$ 1.025 & 0.493 $\pm$ 0.115 \\
      & Adam Reset & \textcolor{OliveGreen}{23.344 $\pm$ 2.075} & 8.110 $\pm$ 1.559 & 8.424 $\pm$ 1.650 & 0.703 $\pm$ 0.088 \\
      & MSGD & 22.160 $\pm$ 1.123 & 8.735 $\pm$ 0.963 & 8.196 $\pm$ 0.822 & 0.547 $\pm$ 0.086 \\
      & MSGD Reset & \textcolor{Blue}{22.628 $\pm$ 1.500} & \textcolor{Blue}{10.345 $\pm$ 1.939} & \textcolor{Blue}{9.536 $\pm$ 1.589} & \textcolor{Blue}{0.493 $\pm$ 0.100} \\
    \midrule
    \multirow[c]{6}{*}{\shortstack{Split\\mini-ImageNet\\{\footnotesize \textbf{(5K buffer)}}}}
      & NGM-SGD (ours) & \textbf{41.004 $\pm$ 2.139} & \textbf{20.435 $\pm$ 2.995} & \textbf{22.016 $\pm$ 2.517} & 0.541 $\pm$ 0.071 \\
      & SGD & 38.008 $\pm$ 2.553 & 19.480 $\pm$ 3.139 & 21.608 $\pm$ 2.699 & \textbf{0.490 $\pm$ 0.090} \\
      & Adam & 39.916 $\pm$ 2.115 & 15.625 $\pm$ 2.222 & 18.744 $\pm$ 1.973 & 0.621 $\pm$ 0.054 \\
      & Adam Reset & \textcolor{OliveGreen}{40.812 $\pm$ 2.082} & 11.295 $\pm$ 3.039 & 15.228 $\pm$ 2.737 & 0.748 $\pm$ 0.069 \\
      & MSGD & 38.404 $\pm$ 1.924 & 15.865 $\pm$ 3.251 & 18.440 $\pm$ 2.669 & 0.589 $\pm$ 0.090 \\
      & MSGD Reset & \textcolor{Blue}{38.748 $\pm$ 2.121} & 14.565 $\pm$ 1.951 & 17.284 $\pm$ 1.832 & 0.618 $\pm$ 0.054 \\
    \midrule
    \multirow[c]{6}{*}{\shortstack{Rotated\\MNIST\\{\footnotesize \textbf{(2K buffer)}}}}
      & NGM-SGD (ours)  & \textbf{92.974 $\pm$ 0.221} & \textbf{87.731 $\pm$ 1.722} & \textbf{88.981 $\pm$ 1.160} & \textbf{0.064 $\pm$ 0.019} \\
      & SGD & 91.986 $\pm$ 0.240 & 83.096 $\pm$ 3.617 & 85.690 $\pm$ 2.421 & 0.103 $\pm$ 0.040 \\
      & Adam & 92.956 $\pm$ 0.263 & 82.554 $\pm$ 1.692 & 85.415 $\pm$ 1.142 & 0.121 $\pm$ 0.019 \\
      & Adam Reset & 92.655 $\pm$ 0.271 & \textcolor{OliveGreen}{86.311 $\pm$ 1.033} & \textcolor{OliveGreen}{87.986 $\pm$ 0.708} & \textcolor{OliveGreen}{0.075 $\pm$ 0.012} \\
      & MSGD & 92.597 $\pm$ 0.363 & 80.860 $\pm$ 1.307 & 84.255 $\pm$ 0.887 & 0.135 $\pm$ 0.015 \\
      & MSGD Reset & \textcolor{Blue}{92.887 $\pm$ 0.347} & \textcolor{Blue}{81.000 $\pm$ 2.222} & \textcolor{Blue}{84.540 $\pm$ 1.494} & 0.135 $\pm$ 0.025 \\
    \midrule
    \multirow[c]{6}{*}{\shortstack{Domain\\CIFAR-100\\{\footnotesize \textbf{(1K buffer)}}}}
      & NGM-SGD (ours) & 51.220 $\pm$ 0.972 & 39.255 $\pm$ 2.888 & 40.150 $\pm$ 1.942 & 0.262 $\pm$ 0.070 \\
      & SGD & 48.220 $\pm$ 1.509 & 32.165 $\pm$ 2.792 & 34.947 $\pm$ 1.930 & 0.363 $\pm$ 0.074 \\
      & Adam & \textbf{53.127 $\pm$ 0.647} & 41.140 $\pm$ 4.926 & 42.423 $\pm$ 3.305 & 0.248 $\pm$ 0.083 \\
      & Adam Reset & 52.743 $\pm$ 1.112 & \textcolor{OliveGreen}{\textbf{41.930 $\pm$ 2.188}} & \textcolor{OliveGreen}{\textbf{42.923 $\pm$ 1.594}} & \textcolor{OliveGreen}{\textbf{0.212 $\pm$ 0.051}} \\
      & MSGD & 50.150 $\pm$ 0.916 & 37.045 $\pm$ 2.477 & 38.763 $\pm$ 1.697 & 0.282 $\pm$ 0.060 \\
      & MSGD Reset & \textcolor{Blue}{51.060 $\pm$ 1.036} & \textcolor{Blue}{39.205 $\pm$ 2.454} & \textcolor{Blue}{40.387 $\pm$ 1.707} & \textcolor{Blue}{0.254 $\pm$ 0.061} \\
    \midrule
    \multirow[c]{6}{*}{\shortstack{Domain\\CIFAR-100\\{\footnotesize \textbf{(5K buffer)}}}}
      & NGM-SGD (ours) & 59.927 $\pm$ 0.934 & 43.110 $\pm$ 3.269 & 46.497 $\pm$ 2.209 & 0.316 $\pm$ 0.055 \\
      & SGD & 56.317 $\pm$ 0.781 & 39.660 $\pm$ 3.946 & 43.177 $\pm$ 2.647 & 0.332 $\pm$ 0.066 \\
      & Adam & 60.287 $\pm$ 1.155 & 44.040 $\pm$ 3.110 & 47.767 $\pm$ 2.154 & \textbf{0.296 $\pm$ 0.053} \\
      & Adam Reset & \textcolor{OliveGreen}{\textbf{61.577 $\pm$ 0.911}} & \textcolor{OliveGreen}{\textbf{44.120 $\pm$ 2.349}} & \textcolor{OliveGreen}{\textbf{47.783 $\pm$ 1.650}} & 0.310 $\pm$ 0.038 \\
      & MSGD & 58.373 $\pm$ 0.910 & 39.830 $\pm$ 2.043 & 44.260 $\pm$ 1.414 & 0.343 $\pm$ 0.036 \\
      & MSGD Reset & 57.213 $\pm$ 1.207 & \textcolor{Blue}{41.370 $\pm$ 3.691} & \textcolor{Blue}{44.790 $\pm$ 2.513} & \textcolor{Blue}{0.307 $\pm$ 0.064} \\
    \bottomrule
  \end{tabular}
  }
  \end{adjustbox}
\end{table*}

\begin{figure*}[pos=!htbp]
  \centering
  \includegraphics[width=\textwidth]{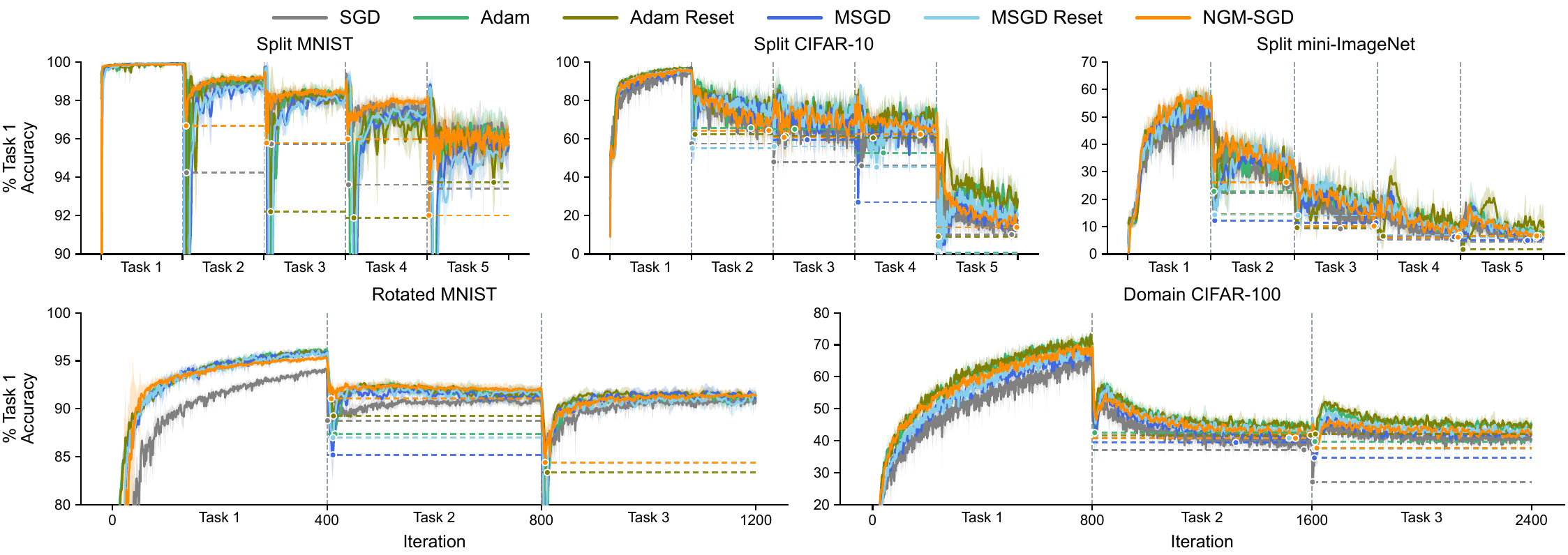}
  \caption{\textbf{Task 1 accuracy across replay experiments with a 1K buffer.} Each panel shows per-iteration accuracy at task switches for all optimizers under a limited replay memory. On simple benchmarks (Split MNIST and Rotated MNIST), the stability gap is visible, whereas on more complex datasets (Split CIFAR-10, Split mini-ImageNet, and Domain CIFAR-100), the small buffer induces substantial steady-state forgetting that dominates over the transient drop. This illustrates how poor approximations of the joint loss can obscure the stability gap we aim to study.}
  \label{fig:app_ER_testAcc}
\end{figure*}

However, when the replay buffer is very limited (e.g. 1K samples), especially on harder benchmarks such as Split CIFAR-10, Split mini-ImageNet, and Domain CIFAR-100, stability gap measurements become entangled with steady-state forgetting, which can obscure the phenomenon we aim to isolate (Figure~\ref{fig:app_ER_testAcc}). With insufficient capacity to store representative coverage of past tasks, the optimizer converges to genuinely suboptimal solutions on older tasks, leading to a long-term loss of performance that can exceed the transient drop induced by the stability gap itself. This behavior aligns with our rationale for studying the stability gap in isolation: the transient form of forgetting should be distinguished from the steady-state forgetting that arises from poor approximations of the joint loss (\emph{catastrophic forgetting}, Figure~\ref{fig:stability_gap_explainer}). For this reason, we also ran experiments with larger buffers, enabling models to recover performance on earlier tasks after each switch and revealing the stability gap in a cleaner, more interpretable manner.

Overall, ER experiments confirm that NGM-SGD continues to attenuate the stability gap even in a more realistic CL scenario where replay partially mitigates catastrophic forgetting, although in Domain CIFAR-100 Adam achieves higher performance metrics, potentially because it converges to a more favorable minimum. Across the benchmarks reported in Table~\ref{tab:ER_metrics}, our method generally shows smaller stability drops and higher minimum accuracies than the other optimizer baselines, indicating that neuromodulatory gain dynamics remain effective beyond the joint training setting. The reset variants of MSGD and Adam exhibit mixed behavior across benchmarks, sometimes reducing stability gaps and sometimes increasing them, indicating that resetting the optimizer state at task boundaries can lead to unstable optimization under replay. Moreover, these methods rely on perfect knowledge of task boundaries and explicitly clear their internal state, so they should be viewed as oracle controls rather than realistic CL baselines.
\section{Stability gap and layer freezing}
\label{app:layer_freezing}
Although the locus coeruleus projects broadly throughout the cortical mantle, including sensory cortices~\citep{samuels_szabadi_lc_2008,aston-jones_integrative_2005}, restricting gain modulation to the classifier or final hidden layers is an intentional functional abstraction rather than a literal anatomical mapping. Cortical and noradrenergic gain are classically described as changes in neuronal responsiveness or input-output sensitivity, often without altering stimulus selectivity~\citep{aston-jones_integrative_2005,ferguson_mechanisms_2020}. In convolutional networks, convolutional layers learn a hierarchy of sensory features, whereas later layers map these representations onto task outputs~\citep{lecun_gradient-based_1998,goodfellow_DL2016}. Applying adaptive gain throughout the backbone during learning could therefore modify the feature basis itself, conflating sensitivity modulation with representational plasticity. We consequently apply gain downstream of the learned feature representation, where it can regulate readout sensitivity and decision dynamics without directly overwriting the sensory feature basis~\citep{aston-jones_integrative_2005,ferguson_mechanisms_2020}.
\begin{figure}[pos=!htbp]
  \centering
  \includegraphics[width=0.9\columnwidth]{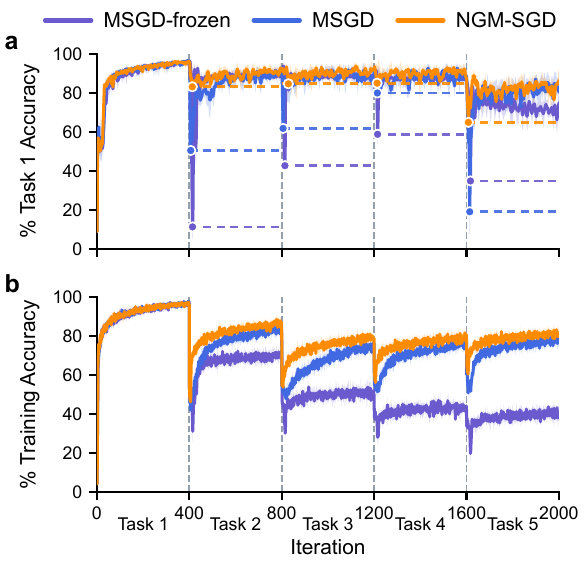}
    \caption{\textbf{Effect of backbone freezing on the stability gap in Split CIFAR-10.} \textbf{a)} Task 1 test accuracy across five sequential tasks for NGM-SGD (Algorithm~\ref{alg:ngm_sgd}; orange), momentum-SGD (MSGD; blue), and MSGD with the convolutional backbone frozen after Task 1 (MSGD-frozen; cyan). Dashed horizontal segments and circular markers indicate the minimum Task 1 accuracy within each task. \textbf{b)} Training accuracy across tasks. Curves and shaded regions show the mean $\pm$ standard deviation across seeds.}
  \label{fig:frozen_backbone}
\end{figure}
To determine whether stabilizing the feature extractor alone could produce a similar effect, we trained an additional Split CIFAR-10 baseline in which the convolutional backbone was frozen after Task 1 and only the classification head remained trainable (Figure~\ref{fig:frozen_backbone}). This intervention does not consistently reduce the stability gap in Task 1 accuracy and leads to substantially lower training accuracy on subsequent tasks. This is consistent with evidence that the stability gap arises predominantly from changes in the classification head rather than the backbone~\citep{lapacz_exploring_2024}: freezing early layers leaves the principal source of transient instability trainable while simultaneously restricting the model's capacity to adapt. NGM-SGD instead acts directly on the downstream layers implicated in the stability gap while preserving their ability to learn the new tasks.





\bibliographystyle{cas-model2-names}

\bibliography{cas-refs}



\end{document}